\documentclass[lettersize,journal]{IEEEtran}
\usepackage{amsmath,amsfonts}
\usepackage{algorithm}
\usepackage{array}
\usepackage{textcomp}
\usepackage{stfloats}
\usepackage{url}
\usepackage{verbatim}
\usepackage{graphicx}
\usepackage{cite}
\usepackage{bm}
\usepackage{array}
\usepackage{amsmath,amssymb}
\usepackage{booktabs}
\usepackage{array}
\usepackage{multirow}
\usepackage{stmaryrd}
\usepackage{colortbl}
\usepackage{threeparttable}

\usepackage{algpseudocode}
\usepackage{subcaption}
\usepackage{enumitem}

\newcommand{\cparagraph}[1]{\smallskip\noindent\textbf{#1}}

\def\m{\bm{m}}
\def\x{\bm{x}}
\def\z{\bm{z}}

\newcommand{\eg}[0]{\textit{e.g.}}
\newcommand{\etc}[0]{\textit{etc}}
\newcommand{\ie}[0]{\textit{i.e.}}
\newcommand{\VM}[0]{\\[1mm]}

\newcommand{\hlt}[1]{{#1}}

\begin{document}

\title{Mask and Restore: Blind Backdoor Defense at Test Time for Local Attacks}

\author{Tao Sun, Lu Pang, Weimin Lyu, Chao Chen, Haibin Ling
\thanks{T. Sun, L. Pang, W. Lyu, C. Chen, H. Ling are with Department of Computer Science, Stony Brook University, Stony Brook, NY 11794, USA.}
\thanks{Manuscript received April xx, xxxx; revised August xx, xxxx.}}

\markboth{PREPRINT}%
{Sun \MakeLowercase{\textit{et al.}}: }


\maketitle

\begin{abstract}
 Deep neural networks are vulnerable to backdoor attacks, where an adversary manipulates the model behavior through overlaying images with special triggers. Existing backdoor defense methods often require accessing a few validation data and model parameters, which is impractical in many real-world applications, \eg, when the model is provided as a cloud service. In this paper, we address the practical task of blind backdoor defense at test time, in particular for local attacks and black-box models. The true label of every test image needs to be recovered on the fly from a suspicious model regardless of image benignity. We consider test-time image purification that incapacitates local triggers while keeping semantic contents intact. Due to diverse trigger patterns and sizes, the heuristic trigger search can be unscalable. We circumvent such barrier by leveraging the strong reconstruction power of generative models, and propose \emph{Blind Defense with Masked AutoEncoder} (BDMAE). BDMAE detects possible local triggers using image structural similarity and label consistency between the test image and MAE restorations. The detection results are then refined by considering trigger topology. Finally, we fuse MAE restorations adaptively into a purified image for making prediction. Extensive experiments under different backdoor settings validate its effectiveness and generalizability. 
\end{abstract}


\section{Introduction}

\IEEEPARstart{D}{eep} neural networks have been widely used in various computer vision tasks, like image classification~\cite{Krizhevsky2012imagenet,masana2022class}, object detection~\cite{girshick2014rich,wang2021salient}, image segmentation~\cite{minaee2021image}, \etc. Despite the superior performances, their vulnerability to backdoor attacks has raised increasing concerns~\cite{gu2019badnets,nguyen2020IAB,turner2019lc}. During training, an adversary can maliciously inject a small portion of poisoned data. These images contain special triggers that are associated with specific target labels. At inference, the backdoored model behaves normally on clean images but makes incorrect predictions on images with triggers.

Backdoor attacks can be roughly categorized as local attacks and global attacks. Local attacks attach local (and usually visible) triggers to the images, which occupy a small portion of the image region~\cite{gu2019badnets,nguyen2020IAB,turner2019lc}.  Global attacks use global (and usually invisible) triggers that span over the entire image region~\cite{barni2019sig,nguyen2021wanet,liu2020reflection,wang2022bppattack}. The two types of attacks have different properties that need to be handled separately. 
%
%
%
{Local backdoor attacks are generally more effective and practical compared to global backdoor attacks. They embed triggers into specific regions of an image, enhancing their stealthiness and making them easier to execute without detection. These localized triggers can be precisely controlled, ensuring a higher success rate with minimal effort. On the other hand, global backdoor attacks distribute triggers across the entire image, which can dilute their effectiveness and make the attack less stealthy. Moreover, separating trigger regions from content regions in global attacks is challenging, making the attacks more complex and less reliable. Thus, local backdoor attacks offer a superior balance of stealthiness and ease of implementation.}

In this paper, we focus on local attacks as they are easier to apply in the real world. To defend against backdoor behaviors, existing methods often require accessing a few validation data and model parameters. Some works reverse-engineer triggers~\cite{wang2019NC,guan2022shapley}, and mitigate backdoor by pruning bad neurons or retraining models~\cite{liu2018fine-pruning,wang2019NC,zeng2021i-bau}. The clean data with labels they require, however, are often unavailable. A recent work shows that the backdoor behaviors could be cleansed with unlabeled or even out-of-distribution data~\cite{pang2022backdoor}. Instead of modifying the model, Februus~\cite{doan2020februus} detects triggers with GradCAM~\cite{selvaraju2017grad}, and feeds purified images to the model. 



All these defending methods, although effective, assume the model is known. Such white-box assumption, however, may be violated in many real-world scenarios. Due to increasing concerns on data privacy and intellectual property, many models are provided as black-boxes where detailed parameters are concealed~\cite{dong2021black,guoaeva,chen2019deepinspect}, \eg, a cloud service API. It is thus crucial to address the problem for black-box models. 

\begin{figure*}[!t]
\centering	
\includegraphics[width=1.0\linewidth]{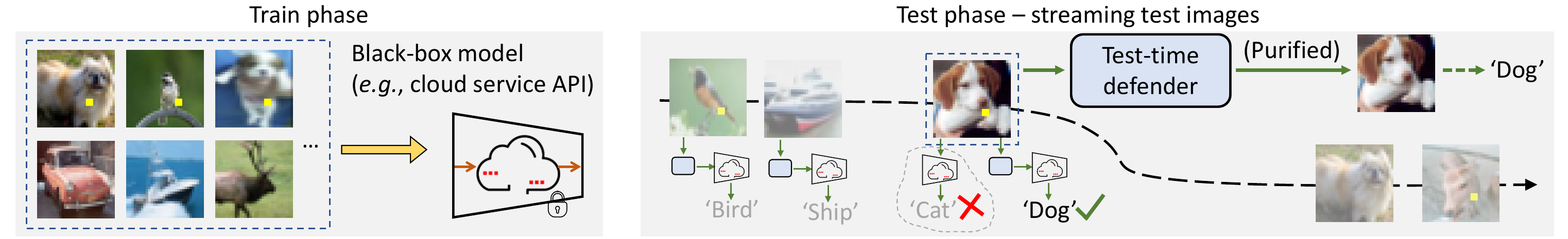}
\vspace{0mm}	
\caption{Test time blind backdoor defense with a black-box prediction model (may be backdoored). Test images come in a stream. The defender purifies them to predict the correct labels on-the-fly. This paper focuses on local visible triggers, which are easy to apply in the real world, and robust against image transformations. Triggers can be complex random patches, social media logos, irregular curves, \textit{etc.}}
\label{fig:task}
\vspace{0mm}
\end{figure*}

In this paper, we tackle the practical yet challenging setting, and address the task of \emph{Blind Backdoor Defense at Test Time}, in particular for local attacks and black-box models. \emph{Blind} means that there is no information on whether the model and test images are backdoored or not. Shown in Fig.~\ref{fig:task}, the prediction model is black-box and may have been injected a backdoor. Test images come in a data stream. The true label of each test image is unknown; it needs to be recovered on the fly only from the hard label predictions of the suspicious model, without accessing additional data. This is a very challenging task that cannot be solved by existing defense methods. Simply applying test-time image transformations~\cite{gao2019strip,sarkar2020backdoor,qiu2021deepsweep} without model retraining compromises accuracies on clean inputs~\cite{sarkar2020facehack}. Heuristic trigger search in image space~\cite{udeshi2022model,xiang2022patchcleanser} does not scale to complex triggers or large image sizes.

To address the challenging task, we resort to the strong reconstruction power of modern image generation models. Intuitively, it can assist us to detect possible triggers and reconstruct the original clean image when the triggers are masked. We propose a novel method called \emph{Blind Defense with Masked AutoEncoder} (BDMAE). Masked Autoencoders~\cite{he2022masked} are scalable self-supervised learners. They randomly mask patches from the input image and reconstruct the missing parts. Each patch corresponds to one of 14$\times$14 tokens. Even using a high masking ratio (\eg, 75\%), the semantic contents can still be recovered. We can therefore search triggers efficiently in the token space. It enables us to generalize to complex trigger patterns or large image sizes.   

Our method belongs to test-time image purification that incapacitates possible local triggers while keeping semantic contents intact. We seek trigger scores that measure how likely each image patch contains triggers. High score regions are then removed and restored with MAE. The whole framework includes three main stages. First, we randomly generate MAE masks, and calculate two types of trigger scores based on image structural similarity and label prediction consistency between test images and MAE restorations, respectively. Then, we use the topology of triggers to refine both scores. The trigger scores help to generate topology-aware MAE masks that cover trigger regions more precisely, and the corresponding MAE restorations in turn help to refine trigger scores. Finally, we fuse multiple MAE restorations from adaptive trigger score thresholding into one purified image, and use that image for label prediction. Our approach is blind to the network architecture, trigger patterns or image benignity. It does not require additional training images for a particular test-time defense task, nor requiring hyper-parameter tuning. Extensive results demonstrate that BDMAE effectively purifies backdoored images without compromising clean images. BDMAE is generalizable to diverse trigger sizes and patterns.

Our main contributions are summarized as follows:
\begin{enumerate}[leftmargin=*,itemsep=-0.5mm]
\vspace{0mm}\item We address the practical task of blind backdoor defense at test time for local attacks and black-box models. Despite some general techniques for simple attacks, this challenging task has not been formally and systematically studied.

\item We propose to leverage generative models to assist backdoor defense. It opens a new door to design general defense methods under limited data using abundant public foundation models. 

\item A novel framework of Blind Defense with Masked AutoEncoders (BDMAE) is devised to detect possible triggers and restore images on the fly. Three key stages are delicately designed to generalize to different defense tasks without tuning hyper-parameters.

\item We evaluate our method on four benchmarks, Cifar10, GTSRB, ImageNet and VGGFace2. Regardless of model architectures, image sizes or trigger patterns, our method obtains superior accuracies on both backdoored images and clean images.
\end{enumerate}

\vspace{0mm}
\section{Related Works}
\vspace{0mm}\textbf{Backdoor attack.}
BadNet~\cite{gu2019badnets} is the earliest work on backdoor attack~\cite{niu2024towards,goldblum2022dataset}. It attaches a checkerboard trigger to images and associates them with specific target labels. Many different trigger patterns are used in later works~\cite{nguyen2020IAB,turner2019lc,wenger2021backdoor}. These triggers are visible local patches in the images. Visible global triggers are used in~\cite{chen2017targeted,barni2019sig}. To make the attack stealthy, invisible patterns~\cite{li2021invisible,zhong2022imperceptible,zhao2022defeat} and attacking strategies based on reflection phenomenon~\cite{liu2020reflection}, image quantization and dithering~\cite{wang2022bppattack}, style transfer~\cite{cheng2021deep} and elastic image warping~\cite{nguyen2021wanet} are proposed. Although these stealthy attacks are less perceptible to humans, they are vulnerable to noise perturbations or image transformations. To make it hard for defenders to reconstruct triggers, sample-specific backdoor attacks~\cite{li2021invisible,nguyen2020IAB} are proposed. This paper focuses on visible triggers of local patches. The triggers can be either shared by samples or sample-specific.

\vspace{0mm}\cparagraph{Backdoor defense.}
Backdoor defense aims to mitigate backdoor behaviors. The training-stage defenses attempt to design robust training mechanism via decoupling training process~\cite{huang2022backdoor}, introducing multiple gradient descent mechanism~\cite{li2021anti} or modifying linearity of trained models~\cite{wang2022training}. However, intruding the training stage is often infeasible. Model reconstruction defenses mitigate backdoor behaviors by pruning bad neurons or retraining models using clean labeled data~\cite{liu2018fine-pruning,wang2019NC,zeng2021i-bau}. A recent work shows that backdoor behaviors could be cleansed by distillation on unlabeled data or even out-of-distribution data~\cite{pang2022backdoor}. Februus~\cite{doan2020februus} is a test-time defense method. It detects triggers with GradCAM~\cite{selvaraju2017grad}, and feeds purified images to the model. 

Recently, black-box backdoor models have drawn increasing attention~\cite{chen2019deepinspect,dong2021black,guoaeva,zhang2021tad}. In this setting, model parameters are concealed for data privacy or intellectual property. 
These works focus on identifying backdoored models, and usually reject predictions for such situations. Differently, we handle the task of blind backdoor defense at test time, aiming to obtain true label of every test image on the fly, with only access to the hard-label predictions. Test-time image transformation~\cite{gao2019strip,sarkar2020backdoor,qiu2021deepsweep} and heuristic trigger search in image space~\cite{udeshi2022model} do not work well. 



\vspace{0mm}\cparagraph{Masked AutoEncoder} (MAE)~\cite{he2022masked} are scalable self-supervised learners based on Vision Transformer~\cite{dosovitskiy2021an}. It masks random patches of the input image, and restore the missing pixels. MAE has been used in many vision tasks~\cite{bachmann2022multimae,pang2022masked,xie2022simmim}. Motivated by the powerful and robust data generation ability, for the first time we leverage MAE to detect triggers and restore images. 

\vspace{0mm}
\section{{Motivation and Intuition}}

Blind backdoor defense at test time aims to obtain correct label prediction for test images on-the-fly regardless the benignity of images and models. To solve this, test-time image purification is a viable solution that incapacitates backdoor triggers within images while keeping semantic contents intact. Some early works apply a global image transformation like blurring or shrinking~\cite{sarkar2020backdoor,qiu2021deepsweep,li2021backdoor}. However, there is often a trade-off in selecting the strength. A stronger transformation is more likely to incapacitate the trigger but at a higher risk of ruining the semantic information. Recently, diffusion model based image purification methods~\cite{nie2022diffusion,wang2022guided} leverage pretrained diffusion models to restore the content, but they highly reply on the image generation quality. When the test data distribution is different from the pretrained data distribution (\textit{e.g.}, different image resolutions), the generated images may appear overall similar to the original test images but still different in the details. This makes it hard for the classifier to predict true labels.

Our motivation is to locate possible local triggers and restore the missing contents simultaneously. Clean regions are kept intact. Thus, model predictions on clean images or clean models are minimally affected. Searching triggers in images can be challenging considering the diversity of trigger patterns and image sizes. Fortunately, with the help of pretrained Masked AutoEncoders (MAE), we can instead search triggers in the token space and use MAE to restore missing parts. 

Fundamentally different from previous defense works:
\begin{itemize}[leftmargin=*,itemsep=0mm]
    \vspace{0mm}\item We care about accuracies on both clean and backdoored images, unlike other defense methods that filter out backdoored images and refuse to make predictions on them.
    \item We leverage pretrained MAE models mainly to assist trigger search, unlike diffusion model-based methods that leverage pretrained generative models to hallucinate the entire image contents. 
\end{itemize}

\begin{figure*}[!t]
	\centering	
	\includegraphics[width=1.0\linewidth]{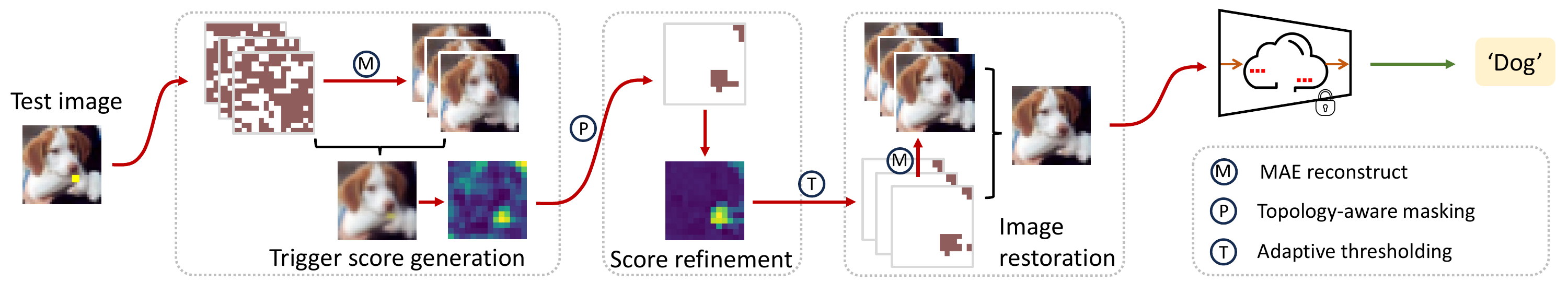}
\vspace{0mm}
	\caption{Framework of our method. For a test image (may or may not be backdoored), we generate the trigger score and refine it by considering the topology of triggers. The purified image obtained from adaptive restoration is used for making prediction.} 
	\label{fig:frame}
\vspace{0mm}
\end{figure*}

\vspace{0mm}
\section{Methodology}

\subsection{Problem Formulation}

\vspace{0mm}We first formulate the backdoor attack and defense problems, then detail the proposed method of \emph{Blind Defense with Masked AutoEncoder} (BDMAE) (Fig.~\ref{fig:frame}). Our key idea is to detect possible triggers with the help of MAE. 


\vspace{0mm}\cparagraph{Backdoor attack.}
Given a set of clean data $D=\{(\x,y)\}$, an adversary generates backdoored data $\tilde{D}=\{(\Phi{(\x)}, \eta(y))|(\x,y)\in D\}$, where $\Phi(\cdot)$ transforms a clean image into a backdoored image and $\eta(\cdot)$ transforms its true label into a target label. In this paper, we consider the popular formulation of $\Phi(\x)=(1-\bm{b}_{\x})\odot \x + \bm{b}_{\x}\odot \bm{\theta}_{\x}$, where $\bm{b}_{\x}$ is a binary mask, $\bm{\theta}_{\x}$ is the backdoor trigger,  $\odot$ denotes the Hadamard product~\cite{dong2021black,hu2021trigger,zheng2021topological}. $\eta(y)$ maps all true labels to one predefined target label. The mask and trigger  may not be the same for different images. While triggers can span over the entire image, we only focus on local visible triggers that occupy a small area of the image, which are easy to apply in the real world, and robust against image transformations. A prediction model $f$ is obtained by training on both clean data and backdoored data. In the situation without backdoor attack, $f$ is obtained from clean data only.

\vspace{0mm}\cparagraph{Black-box test-time defense.}
At test time, the suspicious model $f$ is provided as a black box and only its hard label predictions are accessible. The true label of each test image $\x$ needs to be recovered on the fly, without accessing additional data. To realize this, we seek a purified version $\rho(\x)$ such that $f(\rho(\x))$ generates the correct label prediction. The test process is blind to the model or images, meaning that there is no information on whether $f$ is backdoored and whether $\x$ contains triggers. The goal is to achieve high classification accuracies on both clean and backdoored images. 

\vspace{0mm}
\subsection{Trigger Score Generation}

For clarity, we assume that $f$ is backdoored and the test image $\x$ contains triggers. Our method can directly apply to clean models or clean images (\textit{c.r.} Sec.\ref{sec:clean}). Let $\hat{y}=f(\x)$ be its original label prediction. To infer the trigger mask, one can repeatedly block some particular parts of the image and observe how model predictions change~\cite{udeshi2022model}. However, the search space is huge for a common image size. Even worse, when the trigger is complex (\eg, of irregular shape), the model may still predict the target label when some parts of the trigger remain in the image. These issues make the na\"ive trigger search method infeasible in practice.  
 
We overcome the above-mentioned issues by leveraging the generic Masked AutoEncoders (MAE)~\cite{he2022masked}. In MAE, each of the 14$\times$14 tokens corresponds to a square patch of the image. MAE can recover the image content even when 75\% tokens are masked out. This brings two benefits: 1) we can safely use a high masking ratio to remove triggers without changing the semantic label; and 2) since triggers are irrelevant to the content, they will unlikely present in the MAE restorations. To locate possible triggers, there are two complementary approaches:
the \textbf{image-based} method that compares the \textit{structural similarity} between the original image and MAE restorations, and the \textbf{label-based} method that compares the \textit{consistency of label predictions} on the original image and MAE restorations. 

We use both approaches to obtain an image-based trigger score matrix $S^{(i)}\in[0,1]^{14\times 14}$ and a label-based trigger score matrix $S^{(l)}\in[0,1]^{14\times 14}$. Each element of $S^{(i)}$ or $S^{(l)}$ thus implies how likely the corresponding image patch contains backdoor triggers. Compared with searching in the image space of size $H\times W$, searching in the token space of size $14\times 14$ is much more efficient.



Before going to the method, we first describe how to restore $\x$ given a pre-trained MAE $G$ and a token mask $\m\in \{0,1\}^{14\times 14}$. Define a function $\mathtt{resize}(\z;h,w)$ that resizes a tensor $\z$ to size $h\times w$ by interpolation. As shown in Eq.~\ref{eq:MAE}, $\x$ is first resized to 224$\times$224 requested by MAE. Then we use $G$ to reconstruct the image based on $\m$, and resize it back to $H\times W$. The additional steps aim to remove interpolation errors in the unmasked regions from the restoration $\tilde{\x}$.
\begin{equation}\label{eq:MAE}
	\begin{aligned}
		  \bar{\x}    &=\mathtt{resize}\big(G(\mathtt{resize}(\x;224,224);\m);H,W\big) \\
		\tilde{\m}  &=\mathtt{resize}(\m;H,W) \\
		\tilde{\x}  &= \x\odot (1-\tilde{\m})+ \bar{\x}\odot \tilde{\m} \\
		\tilde{G}(\x,\m) &\triangleq (\tilde{\x},\tilde{\m})
	\end{aligned}
\end{equation}
%
%
Now we describe how to obtain trigger scores $S^{(i)}$ and $S^{(l)}$ from MAE restorations. 
Let $\hat{y}=f(\x)$ be its original hard-label prediction. We repeat the following procedure for $N_o$ times indexed by $o$. For each iteration, $N_i$ random token masks $\{\m_{o,i}\in\{0,1\}^{14\times 14}$\} are sampled using a default masking ratio of 75\%. The corresponding MAE reconstructions $\{\tilde{\x}_{o,i}\}$ and masks $\{\tilde{\m}_{o,i}\}$ are extracted from $ \tilde{G}(\x,\m_{o,i})$ based on Eq.~\ref{eq:MAE}. Their hard-label predictions are $\{\hat{y}_{o,i}=f(\tilde{\x}_{o,i})\}$.

\vspace{0mm}\cparagraph{Image-based score $S^{(i)}$.} We fuse $N_i$ restorations into one image $\tilde{\x}_o$ by:
\begin{equation}\label{eq:fuse}
	\tilde{\x}_o=\mathcal{F}\big(\{\tilde{\x}_{o,i}\},\{\tilde{\m}_{o,i}\}\big)=\sum_i (\tilde{\x}_{o,i}\odot \tilde{\m}_{o,i})	\oslash \sum_i (\tilde{\m}_{o,i}) 	
\end{equation}
where $\odot$ and $\oslash$ are element-wise product and division. In Eq.~\ref{eq:fuse}, only image patches from MAE restorations are kept while other patches from the original image are discarded. The motivation is that triggers may not always be fully masked out, but we do not want them to appear in $\tilde{\x}_o$. We manipulate the sampling of $\{\m_{o,i}\}$ to guarantee that every image patch can be restored with Eq.~\ref{eq:fuse}.

The image-based score is defined as $S^{(i)}=\sum_o [1-\mathtt{resize}(\mathrm{SSIM}(\x, \tilde{\x}_{o});14,14)]/N_{o}$, averaged over $N_o$ repeated procedures. We use Structural Similarity Index Measure (SSIM)~\cite{wang2004image} to calculate the similarity between $\tilde{\x}_o$ and $\x$. The score lies between $-1$ and $1$. As triggers are irrelevant to contents and unlikely present in $\tilde{\x}_o$, SSIM scores in trigger regions will be low. In contrast, the clean regions will be well restored, leading to high SSIM scores. 

\vspace{0mm}\cparagraph{Label-based score $S^{(l)}$}. We average over token masks that lead to different label predictions. Formally, the label-based score is defined as $S^{(l)}=\sum_{o,i}[\m_{o,i}\times(1-\mathbb{I}[\hat{y}=\hat{y}_{o,i}] )]/(N_o N_i)$, where $\mathbb{I}[\cdot]$ is the indicator function. The inconsistency in label predictions usually implies that triggers have been removed by the masks.

The two types of trigger scores are complementary. $S^{(i)}$ favors large and complex triggers, while $S^{(l)}$ favors small triggers. Using them together adapt to diverse trigger patterns. The detailed procedures can be found in Alg.~\ref{alg:triggerGen}.

\begin{figure}[t]
\begin{algorithm}[H]
	\renewcommand{\algorithmicrequire}{\textbf{Input:}}
	\renewcommand{\algorithmicensure}{\textbf{Output:}}
	\caption{Trigger Score Generation} 
	\label{alg:triggerGen} 
	\begin{algorithmic}[1]
		\Require Prediction model $f$, test image $\x$, generic MAE model $G$, repeated times $N_o$, $N_i$.
		\Ensure Trigger scores $S^{(i)}$, $S^{(l)}$
		\State Get original hard-label prediction $\hat{y}=f(\x)$		
		\For {$o=0$ \textbf{to} $N_o$}
		\For {$i=0$ \textbf{to} $N_i$}
		\State Uniformly sample random token mask $\m_{o,i}$
		\State Get MAE reconstruction ${\tilde{\x}_{o,i}}$ and the corresponding mask ${\tilde{\m}_{o,i}}$ from $\tilde{G}(\x,\m_{o,i})$
		\State Get hard-label prediction $\hat{y}_{o,i}=f(\tilde{\x}_{o,i})$
		\EndFor
		\State Fuse restorations into $\tilde{\x}_{o}=\mathcal{F}(\{\tilde{\x}_{o,i}\},\{\tilde{\m}_{o,i}\})$
		\State Calculate structural similarity $I_{o}=\mathrm{SSIM}(\x, \tilde{\x}_{o})$
		\EndFor
		
		\State $S^{(i)}=\sum_o [1-\mathtt{resize}(I_{o};14,14)]/N_{o}$
		\State $S^{(l)}=\sum_{o,i}[\m_{o,i}\times(1-\mathbb{I}[\hat{y}=\hat{y}_{o,i} ])]/(N_o N_i)$
	\end{algorithmic}
\end{algorithm}
\end{figure}

\begin{figure}[t]
\vspace{0mm}
\begin{algorithm}[H]
	\renewcommand{\algorithmicrequire}{\textbf{Input:}}
	\renewcommand{\algorithmicensure}{\textbf{Output:}}
	\caption{Topology-aware Score Refinement} 
	\label{alg:triggerRefine} 
	\begin{algorithmic}[1]
		\Require Prediction model $f$, test image $\x$, generic MAE model $G$, refinement times $N_r$, initial trigger score $S^{(*)}$, mask $\m_{\rm rf}$ for tokens to be refined, $\beta_0=0.05$.
		\Ensure Refined trigger score $S^{(*)}$.
		\State Get original hard-label prediction $\hat{y}=f(\x)$		
		\For {$r=0$ \textbf{to} $N_r$}
		\State Generate a topology-aware token mask $\m_{r}$
		\State $\bar{\m}_{r}=\m_{\rm rf}-\m_{r}$
		\State Get MAE reconstruction $\tilde{\x}_{r}$ from $\tilde{G}(\x,\m_{r})$
		\State Get hard-label prediction $\hat{y}_{r}=f(\tilde{\x}_{r})$ 
		\State $\beta=(1-2\mathbb{I}[ \hat{y}=\hat{y}_{r} ])\times \beta_0$
		\State $S^{(*)}\leftarrow S^{(*)}+\beta\times(\m_{r}-\bar{\m}_{r})$		
		\EndFor
	\end{algorithmic}
\end{algorithm}
\end{figure}

\begin{figure*}[!t]
	\centering	
	\begin{subfigure}{0.095\textwidth}
		\includegraphics[width=\textwidth]{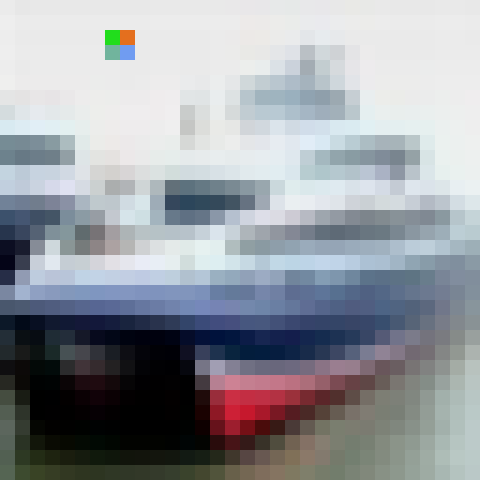}
	\end{subfigure}	
	\begin{subfigure}{0.095\textwidth}
		\includegraphics[width=\textwidth]{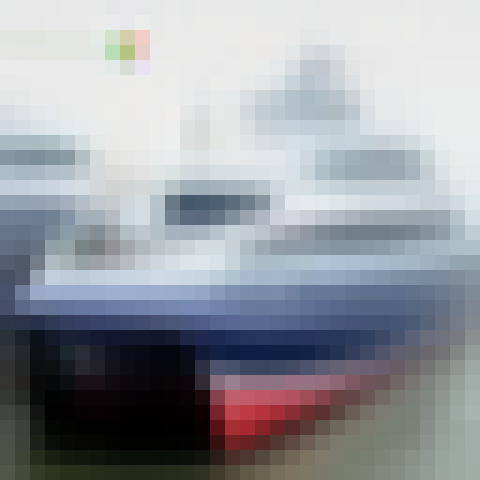}
	\end{subfigure}	
	\begin{subfigure}{0.095\textwidth}
		\includegraphics[width=\textwidth]{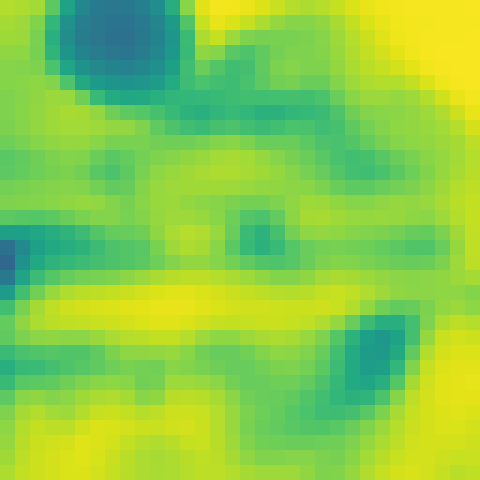}
	\end{subfigure}	
	\begin{subfigure}{0.095\textwidth}
		\includegraphics[width=\textwidth]{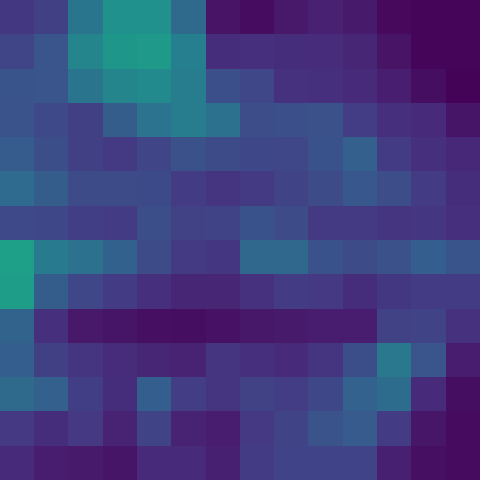}
	\end{subfigure}
	\begin{subfigure}{0.095\textwidth}
		\includegraphics[width=\textwidth]{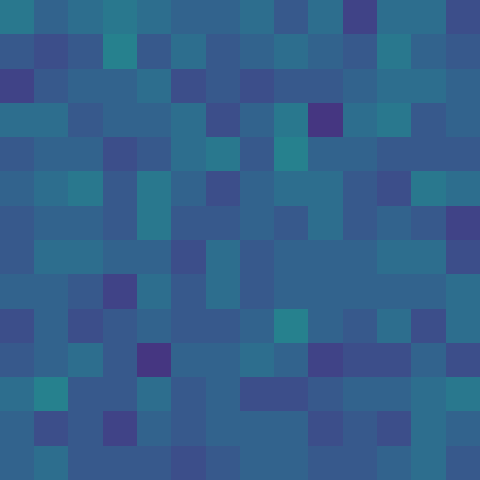}
	\end{subfigure}
	\begin{subfigure}{0.095\textwidth}
		\includegraphics[width=\textwidth]{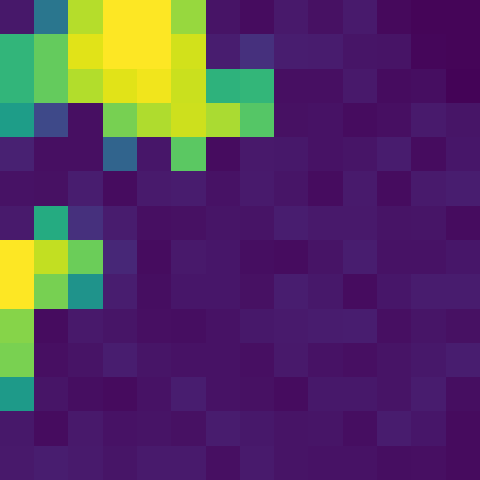}
	\end{subfigure}	
	\begin{subfigure}{0.095\textwidth}
		\includegraphics[width=\textwidth]{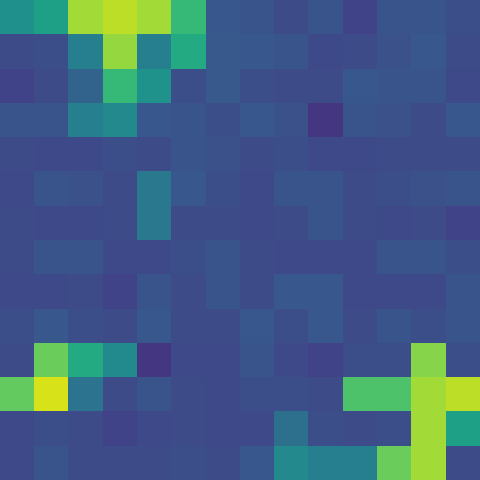}
	\end{subfigure}
	\begin{subfigure}{0.095\textwidth}
		\includegraphics[width=\textwidth]{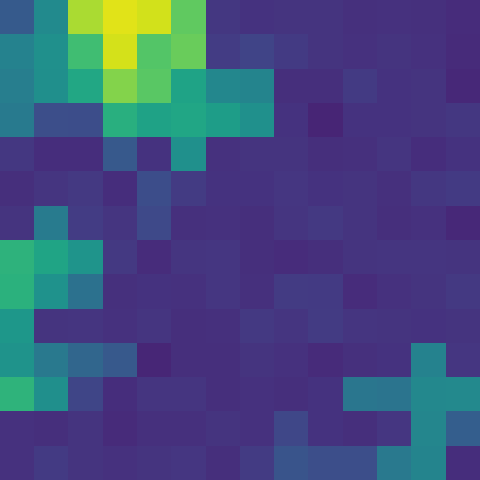}
	\end{subfigure}
	\begin{subfigure}{0.095\textwidth}
		\includegraphics[width=\textwidth]{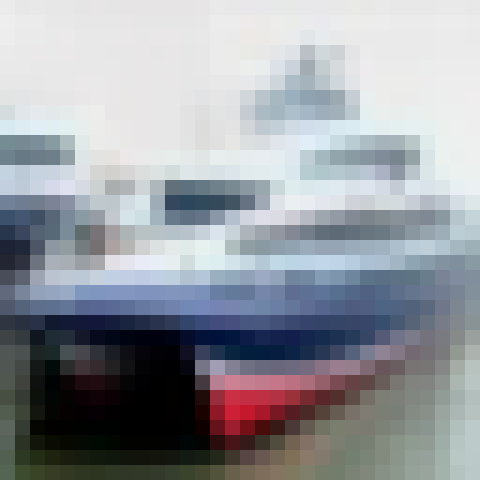}
	\end{subfigure}
        \VM
        
	\begin{subfigure}{0.095\textwidth}
		\includegraphics[width=\textwidth]{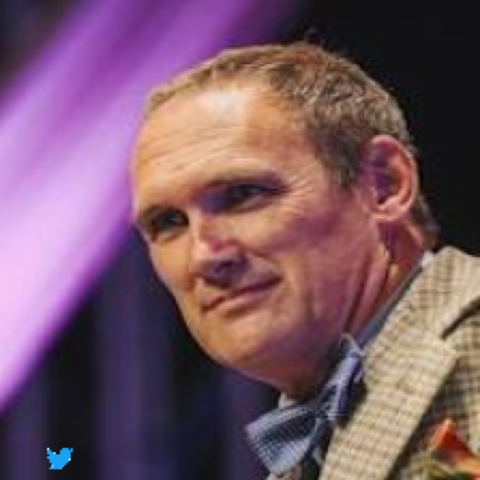}
	\end{subfigure}	
	\begin{subfigure}{0.095\textwidth}
		\includegraphics[width=\textwidth]{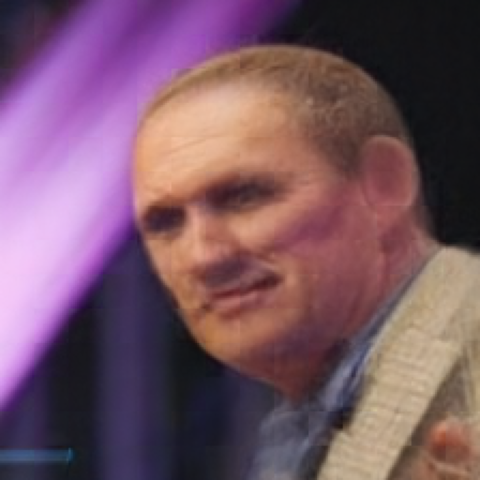}
	\end{subfigure}	
	\begin{subfigure}{0.095\textwidth}
		\includegraphics[width=\textwidth]{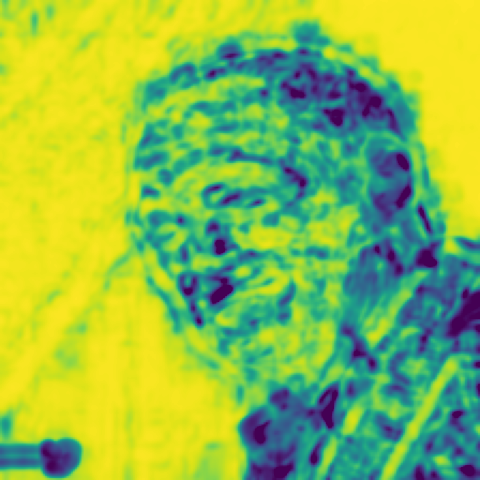}
	\end{subfigure}
	\begin{subfigure}{0.095\textwidth}
		\includegraphics[width=\textwidth]{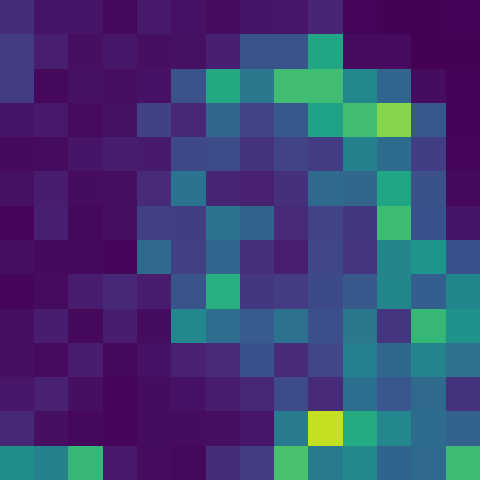}
	\end{subfigure}
	\begin{subfigure}{0.095\textwidth}
		\includegraphics[width=\textwidth]{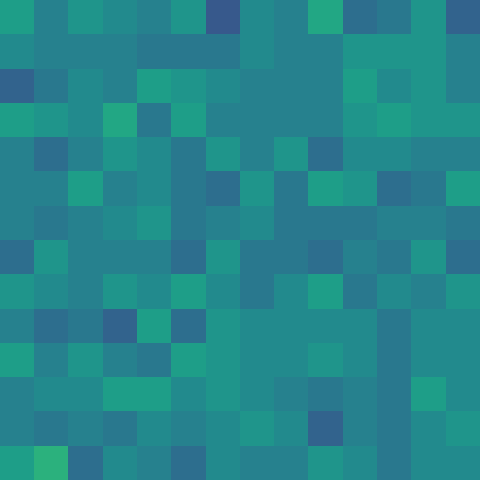}
	\end{subfigure}
	\begin{subfigure}{0.095\textwidth}
		\includegraphics[width=\textwidth]{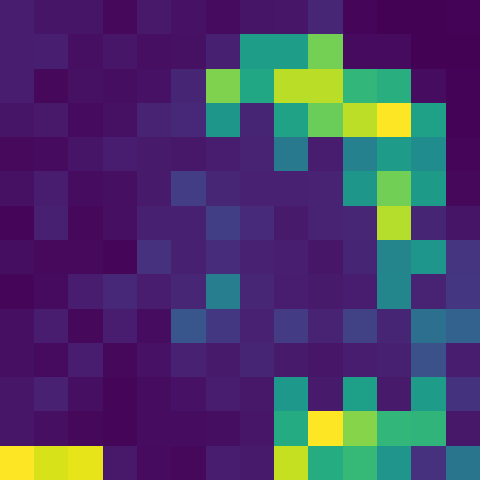}
	\end{subfigure}	
	\begin{subfigure}{0.095\textwidth}
		\includegraphics[width=\textwidth]{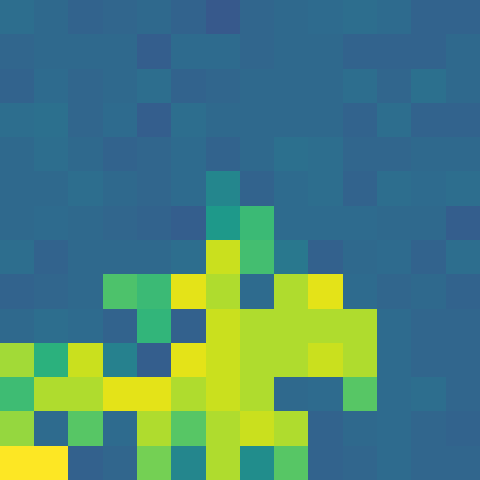}
	\end{subfigure}
	\begin{subfigure}{0.095\textwidth}
		\includegraphics[width=\textwidth]{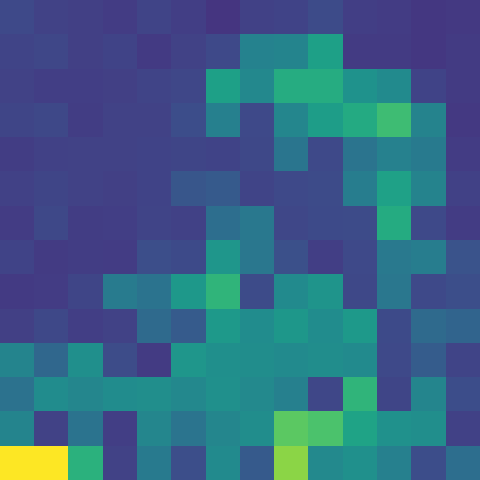}
	\end{subfigure}
	\begin{subfigure}{0.095\textwidth}
		\includegraphics[width=\textwidth]{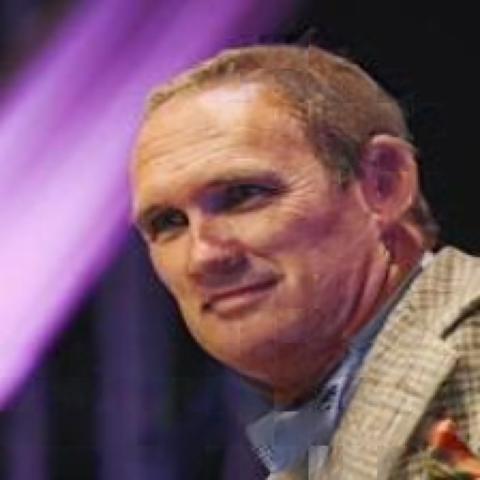}
	\end{subfigure}

	\begin{minipage}{0.095\textwidth}
		\centering
		\scriptsize Original image w/ trigger
	\end{minipage} 
	\begin{minipage}{0.095\textwidth}
		\centering
		\scriptsize Restored image from rand masks 
	\end{minipage} 
	\begin{minipage}{0.095\textwidth}
		\centering
		\scriptsize SSIM score map 
	\end{minipage} 
	\begin{minipage}{0.095\textwidth}
		\centering
		\scriptsize $S^{(i)}$ before refinement  
	\end{minipage} 
	\begin{minipage}{0.095\textwidth}
		\centering
		\scriptsize $S^{(l)}$ before refinement 
	\end{minipage} 
	\begin{minipage}{0.095\textwidth}
		\centering
		\scriptsize $S^{(i)}$ after refinement   
	\end{minipage} 
	\begin{minipage}{0.095\textwidth}
		\centering
		\scriptsize $S^{(l)}$ after refinement   
	\end{minipage} 
	\begin{minipage}{0.095\textwidth}
		\centering
		\scriptsize $S$ for final restoration  
	\end{minipage} 
	\begin{minipage}{0.095\textwidth}
		\centering
		\scriptsize Purified image for prediction  
	\end{minipage} 
\vspace{0mm}
\caption{Sampled visualizations of the defense process. Top: \texttt{Cifar10} with 2$\times$2-color trigger. Bottom: \texttt{VGGFace2} with \emph{twitter} trigger. All the scores are clipped to a range of [0,1], with yellow for high values and blue for low values.}\label{fig:visualization_I}
\vspace{2mm}
\end{figure*}

\begin{figure*}[!t]
	\centering

	\begin{subfigure}{0.077\textwidth}
	\includegraphics[width=\textwidth]{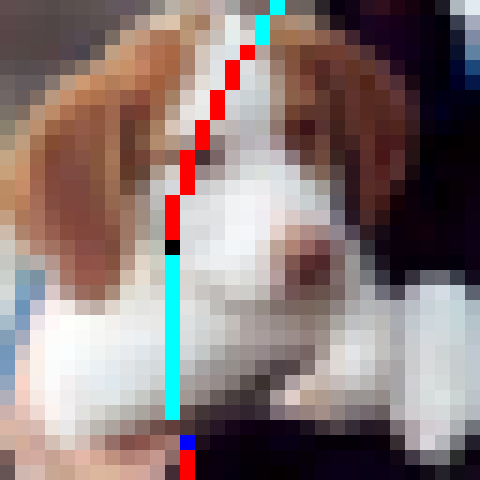}
	\end{subfigure}
	\begin{subfigure}{0.077\textwidth}
		\includegraphics[width=\textwidth]{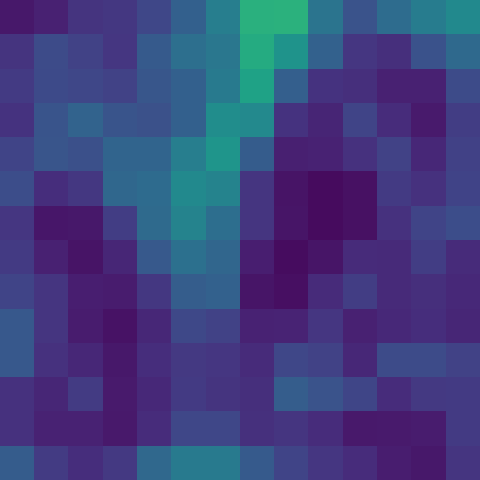}
	\end{subfigure}
	\begin{subfigure}{0.077\textwidth}
		\includegraphics[width=\textwidth]{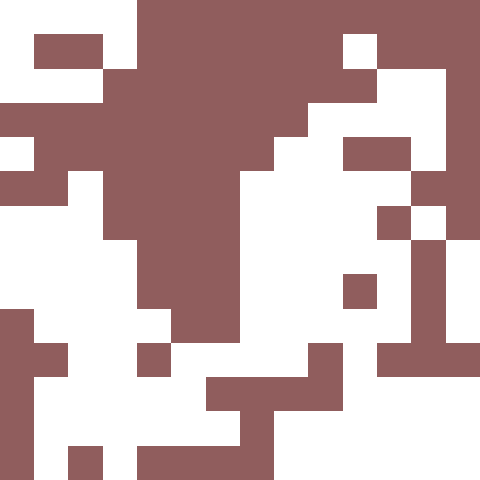}
	\end{subfigure}
	\begin{subfigure}{0.077\textwidth}
		\includegraphics[width=\textwidth]{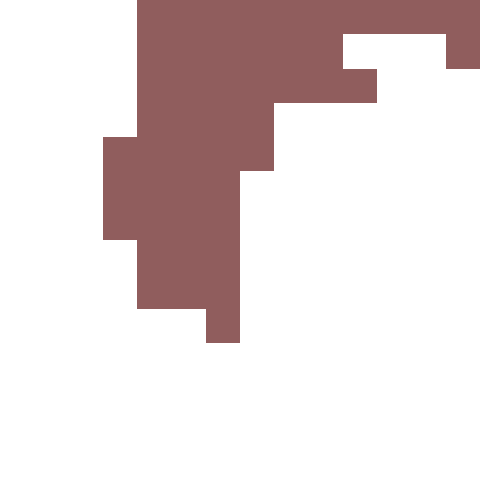}
	\end{subfigure}
	\begin{subfigure}{0.077\textwidth}
		\includegraphics[width=\textwidth]{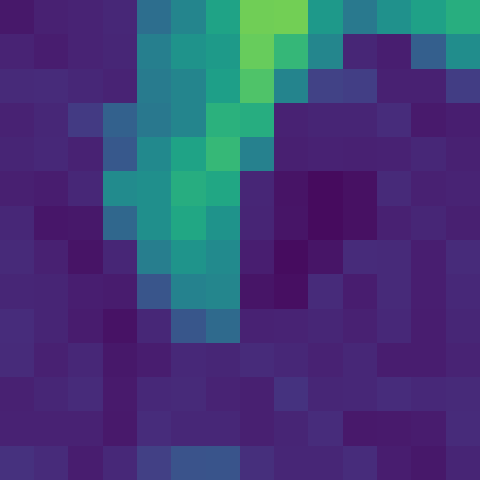}
	\end{subfigure}
	\begin{subfigure}{0.077\textwidth}
		\includegraphics[width=\textwidth]{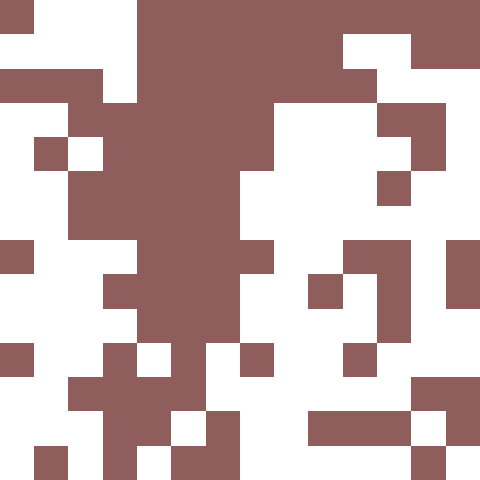}
	\end{subfigure}
	\begin{subfigure}{0.077\textwidth}
		\includegraphics[width=\textwidth]{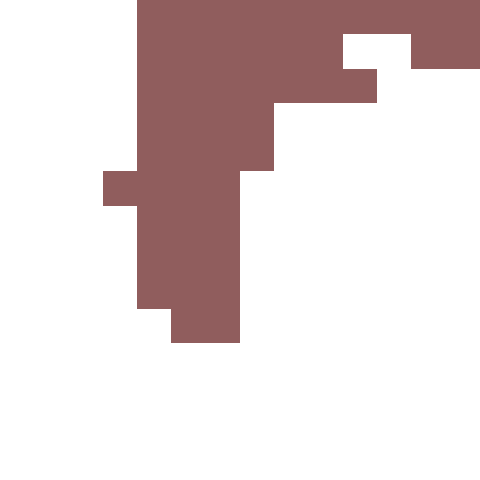}
	\end{subfigure}
	\begin{subfigure}{0.077\textwidth}
		\includegraphics[width=\textwidth]{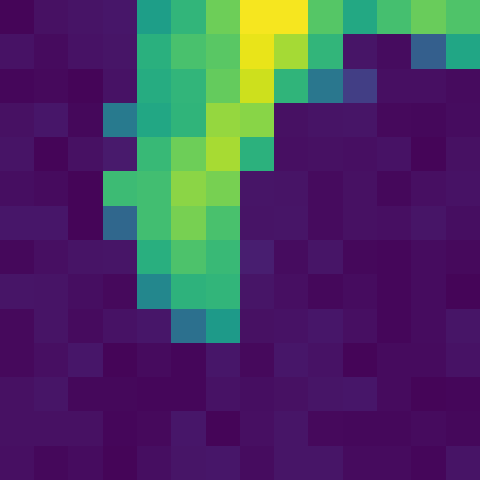}
	\end{subfigure}
	\begin{subfigure}{0.077\textwidth}
		\includegraphics[width=\textwidth]{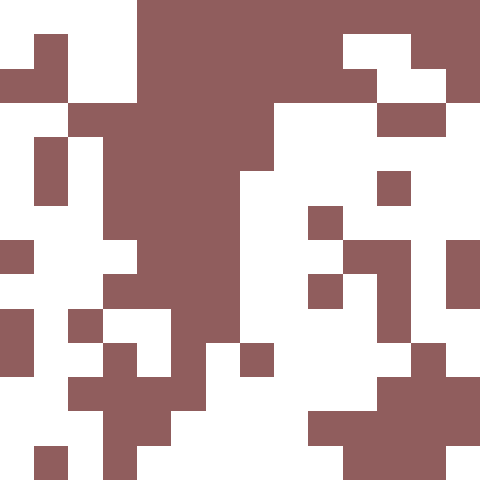}
	\end{subfigure}
	\begin{subfigure}{0.077\textwidth}
		\includegraphics[width=\textwidth]{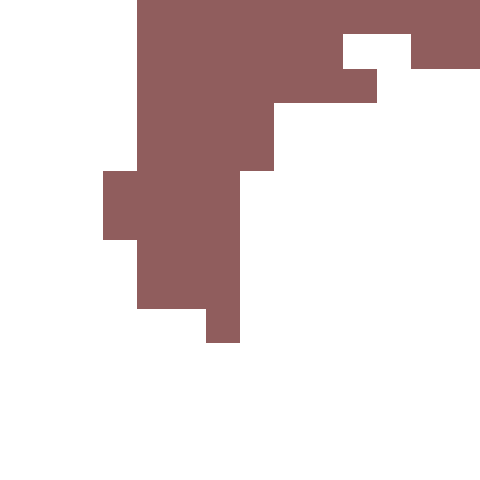}
	\end{subfigure}
	\begin{subfigure}{0.077\textwidth}
		\includegraphics[width=\textwidth]{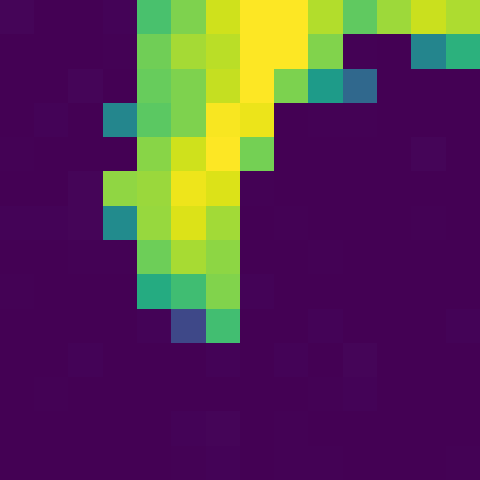}
	\end{subfigure}
	\begin{subfigure}{0.077\textwidth}
		\includegraphics[width=\textwidth]{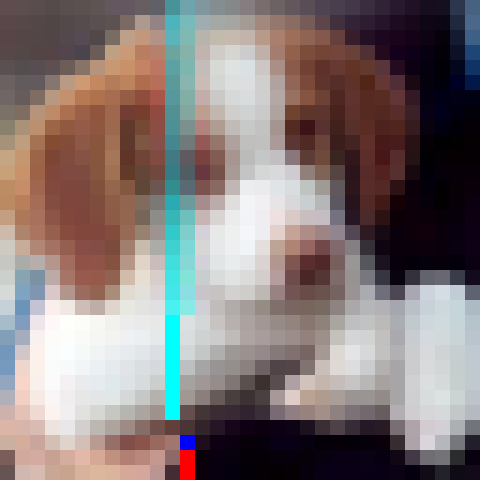}
	\end{subfigure}
    \VM
   
	\begin{subfigure}{0.077\textwidth}  
  \includegraphics[width=\textwidth]{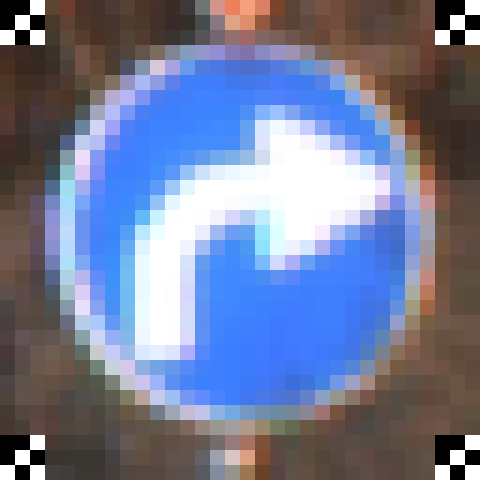}
	\end{subfigure}
	\begin{subfigure}{0.077\textwidth}
		\includegraphics[width=\textwidth]{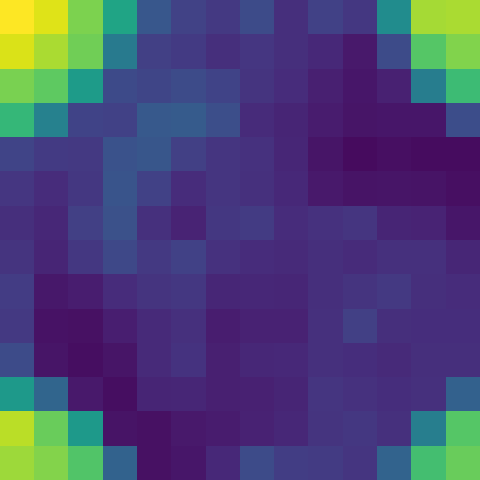}
	\end{subfigure}
	\begin{subfigure}{0.077\textwidth}
		\includegraphics[width=\textwidth]{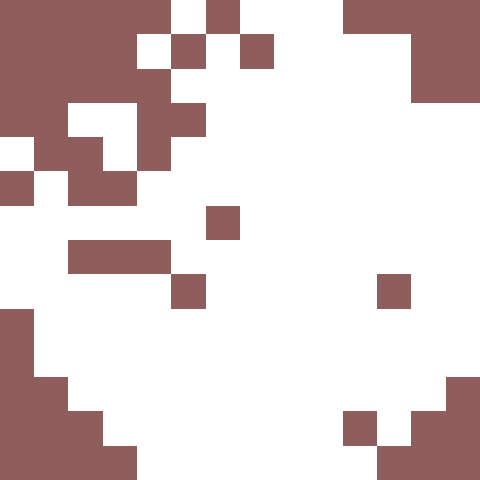}
	\end{subfigure}
	\begin{subfigure}{0.077\textwidth}
		\includegraphics[width=\textwidth]{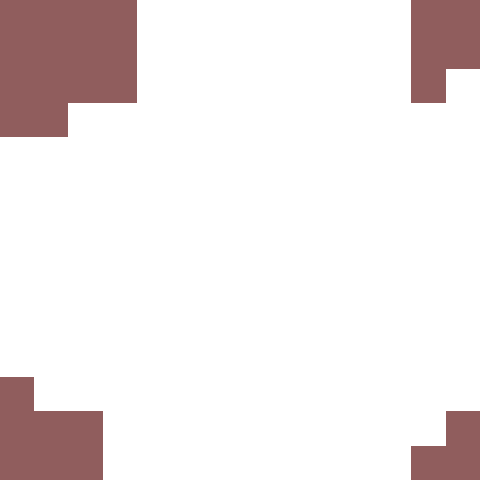}
	\end{subfigure}
	\begin{subfigure}{0.077\textwidth}
		\includegraphics[width=\textwidth]{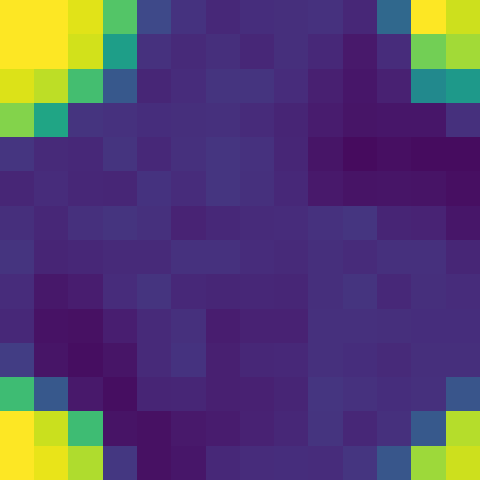}
	\end{subfigure}
	\begin{subfigure}{0.077\textwidth}
		\includegraphics[width=\textwidth]{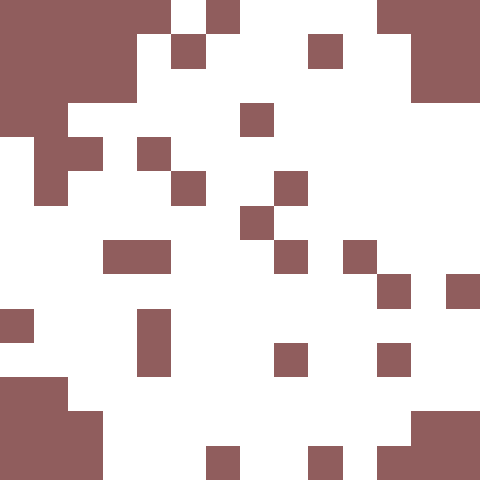}
	\end{subfigure}
	\begin{subfigure}{0.077\textwidth}
		\includegraphics[width=\textwidth]{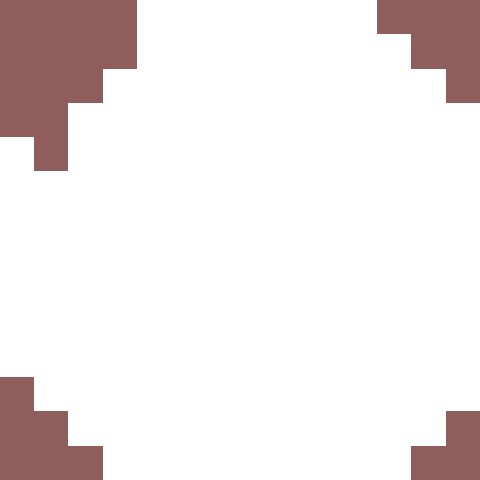}
	\end{subfigure}
	\begin{subfigure}{0.077\textwidth}
		\includegraphics[width=\textwidth]{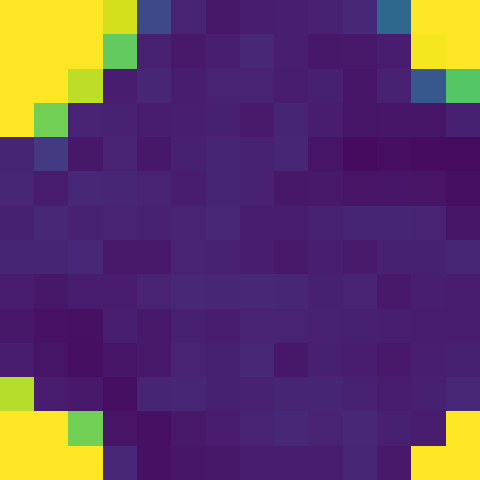}
	\end{subfigure}
	\begin{subfigure}{0.077\textwidth}
		\includegraphics[width=\textwidth]{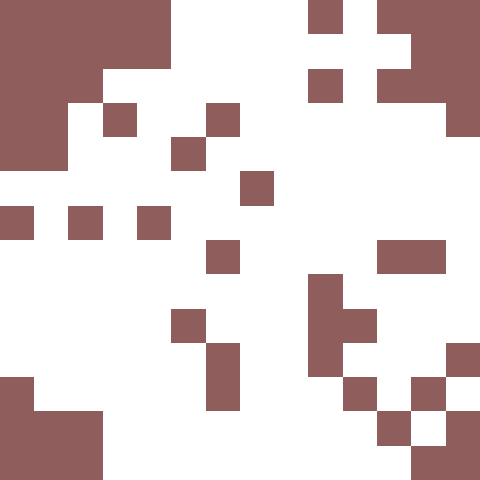}
	\end{subfigure}
	\begin{subfigure}{0.077\textwidth}
		\includegraphics[width=\textwidth]{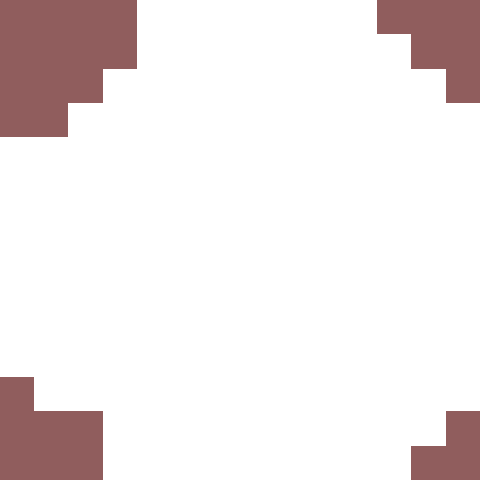}
	\end{subfigure}
	\begin{subfigure}{0.077\textwidth}
		\includegraphics[width=\textwidth]{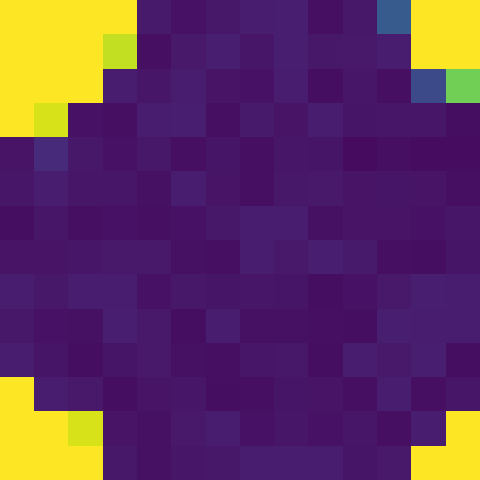}
	\end{subfigure}
	\begin{subfigure}{0.077\textwidth}
		\includegraphics[width=\textwidth]{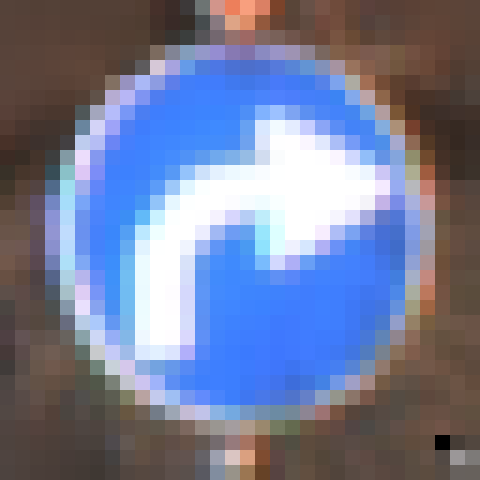}
	\end{subfigure}

	\begin{minipage}{0.077\textwidth}
		\centering
		\scriptsize Original image  
	\end{minipage} 
	\begin{minipage}{0.077\textwidth}
		\centering
		\scriptsize $S^{(i)}$ (0) 
	\end{minipage} 	
	\begin{minipage}{0.077\textwidth}
		\centering
		\scriptsize  $\m_{\rm rf}$ (3)
	\end{minipage} 
	\begin{minipage}{0.077\textwidth}
		\centering
		\scriptsize  $\m_{r}$ (3)
	\end{minipage} 
	\begin{minipage}{0.077\textwidth}
		\centering
		\scriptsize $S^{(i)}$ (3) 
	\end{minipage} 	
	\begin{minipage}{0.077\textwidth}
		\centering
		\scriptsize  $\m_{\rm rf}$ (7)
	\end{minipage} 
	\begin{minipage}{0.077\textwidth}
		\centering
		\scriptsize  $\m_{r}$ (7)
	\end{minipage} 
	\begin{minipage}{0.077\textwidth}
		\centering
		\scriptsize $S^{(i)}$ (7) 
	\end{minipage} 	
	\begin{minipage}{0.077\textwidth}
		\centering
		\scriptsize  $\m_{\rm rf}$ (10)
	\end{minipage} 
	\begin{minipage}{0.077\textwidth}
		\centering
		\scriptsize  $\m_{r}$ (10)
	\end{minipage} 
	\begin{minipage}{0.077\textwidth}
		\centering
		\scriptsize $S^{(i)}$ (10) 
	\end{minipage} 	
	\begin{minipage}{0.077\textwidth}
		\centering
		\scriptsize Purified image  
	\end{minipage} 
\vspace{0mm}
\caption{Visualizations of topology-aware score refinement. Top: \texttt{Cifar10} with IAB. Bottom: \texttt{GTSRB} with LC. The numbers in brackets indicate steps of refinement.}
\label{fig:topology}
\vspace{0mm}
\end{figure*}

\subsection{Topology-aware Score Refinement}
\vspace{0mm}

The trigger scores $S^{(i)}$ and $S^{(l)}$ obtained previously have high values for trigger regions. Nevertheless, they are still very noisy. The difference between scores of trigger regions and clean regions is also small, making it hard to determine a universal threshold for filtering out trigger regions.

We utilize the topology of triggers to refine trigger scores. Note that backdoor triggers are commonly continuous patterns~\cite{hu2021trigger}. The obtained trigger scores indicate possible positions of triggers among the image. With the information, we can generate topology-aware MAE masks $\{\m_r\in\{0,1\}^{14\times 14}\}$ that cover trigger regions more precisely than uniformly sampled ones. This in turn guides us to enhance the difference between score values of clean regions and trigger regions. One issue is that if we apply refinement for all tokens, we may accidentally increase the score values of clean regions in the situation of clean images or clean models. To avoid this, we only focus on the top $L$ tokens that likely contain triggers, with $L=\sum_{r,c}\mathbb{I}[S^{(i)}_{r,c}\geq 0.2]$ or $L=\sum_{r,c} S^{(l)}_{r,c}$. Equivalently, a meta mask $\m_{\rm rf}\in \{0,1\}^{14\times 14}$ can be defined, whose element is 1 only if the corresponding token belongs to the top $L$ tokens. $\m_{\rm rf}$ thus indicates the regions to be refined. 

We use the same procedure to generate topology-aware MAE mask $\m_r$ for both types of trigger scores. The main idea is to sequentially select tokens that have higher trigger scores or are adjacent to already selected tokens. For clarity, let $S^{(*)}$ denote either $S^{(i)}$ or $S^{(l)}$. We initialize $\mathcal{T}=\{t_0\}$ with token $t_0=\arg\max_{t_k} S^{(*)}[t_k]$. Then we repeatedly add token $t_i=\arg\max_{t_k} (S^{(*)}[t_k]+0.5\mathbb{I}[t_k \in \texttt{Adj}(\mathcal{T}) ])\cdot \sigma_{k} $ to $\mathcal{T}$, where $\texttt{Adj}(\mathcal{T})$ includes all 4-nearest neighbors of tokens in $\mathcal{T}$ and $\sigma_{k}\sim U(0,1)$ is a random variable. This step achieves a balance between random exploration and topology-aware exploitation. The process continues until $|\mathcal{T}|=L/2$. The final $\mathcal{T}$ can be converted into an MAE mask $\m_r$, with its complementary part $\bar{\m}_{r}=\m_{\rm rf}-\m_{r}$. 

To refine the trigger score, we obtain the hard-label prediction $\hat{y}_r$ of MAE restoration based on $\m_r$. If $\hat{y}_r\neq \hat{y}$, we increase the score values of 
$S^{(*)}$ by a constant $\beta_0$ for tokens masked by $\m_r$ and $-\beta_0$ for other tokens; otherwise, we modify $S^{(*)}$ in an opposite way. Mathematically, $S^{(*)}\leftarrow S^{(*)}+(1-2\mathbb{I}[ \hat{y}=\hat{y}_{r} ])\times \beta_0\times(\m_{r}-\bar{\m}_{r})$. Since $\|\m_r\|_0=\|\bar{\m}_{r}\|_0=L/2$, the average value of $S^{(*)}$ remains unchanged, while the contrast between trigger region and clean region are enhanced. The procedure is summarized in Alg.~\ref{alg:triggerRefine}.

\vspace{0mm}
\subsection{Adaptive Image Restoration}

The combined trigger score used for label prediction is simply calculated as $S=(S^{(i)}+S^{(l)})/2$. One can convert $S$ into a binary mask based on some predefined threshold, and make prediction on the corresponding MAE restoration. However, the optimal threshold varies across different attack settings considering the diversity of image resolutions, backdoor attack methods, and trigger sizes.

We propose an adaptive image restoration mechanism to adapt to different attacks and datasets automatically. The idea is to fuse restorations from $K$ adaptive thresholds, $\{\tau_1\geq \tau_2\geq \cdots\geq \tau_K\}$. If $\sum_{r,c}\mathbb{I}[S_{r,c}\geq \tau_K]/(14\times 14) \leq 25\%$ is not satisfied, we repeatedly increase all thresholds by a small amount. The rationale is that trigger regions should not dominate the image. These decreasing thresholds lead to a nest structure. We obtain the corresponding MAE restorations $\{\tilde{\x}_{\tau_k},\tilde{\m}_{\tau_k}=\tilde{G}(\x,\m_{\tau_k})\}$, where $\m_{\tau_k}[r,c] = \mathbb{I}[S[r,c]\geq \tau_k]$, and then fuse them into one purified image $\rho(\x)=\mathcal{F}(\{\tilde{\x}_{\tau_k}\},\{\tilde{\m}_{\tau_k}\})$. The model prediction $f(\rho(\x))$ is used for final evaluation. It should be noted that we use the same predefined $\{\tau_{k}\}$ for all tasks, and do not tune them on each dataset.

\subsection{Generalization to Clean Images and Models}\label{sec:clean}
Until now, we assume that both $f$ and $\x$ are backdoored. In practice, we deal with blind defense, meaning that both models and images can be either backdoored or clean. Our method directly applies to any of these situations. The effectiveness on clean images has been validated by CA metric. For clean models, we include discussions in Sec.~\ref{sec:clean}.

\vspace{0mm}
\section{Experiments}


\subsection{Datasets}

We evaluate our method on the commonly used \texttt{Cifar10}~\cite{krizhevsky2009cifar10}, \texttt{GTSRB}~\cite{stallkamp2012gtsrb}, \texttt{VGGFace2}~\cite{cao2018vggface2}, and three \texttt{ImageNet}~\cite{deng2009imagenet} subsets, including \texttt{ImageNet10}, \texttt{ImageNet50} and \texttt{ImageNet100}.

\begin{itemize}
    \item {\texttt{Cifar10}} is a 10-class classification dataset~\cite{krizhevsky2009cifar10} with image size 32$\times$32. There are 50,000 training images and 10,000 test images.

    \item {\texttt{GTSRB}}~\cite{stallkamp2012gtsrb} consists of 43-class traffic signs images of size 32$\times$32, split into 39,209 training images and 12,630 test images. 

    \item {\texttt{VGGFace2}} is a face recognition dataset~\cite{cao2018vggface2}. We use images from 170 randomly selected classes following~\cite{doan2020februus}, and resize them to 224$\times$224. Face recognition is a critical real-world application where backdoor attack may exist.

    \item {\texttt{ImageNet10}, \texttt{ImageNet50} and \texttt{ImageNet100}} are three subsets of \texttt{ImageNet}~\cite{deng2009imagenet}, resized to 224$\times$224. We created them by selecting the first 10 (50, 100) classes in alphabetical order. Each class has about 1,300 training images and 50 test images.
\end{itemize}




\begin{table*}[!t]

\caption{Comparison with diffusion model based image purification method (500 test images). The upper table is with BadNet, and the lower table is with IAB and Blended attacks.}\label{tab:dp_results}

\renewcommand{\tabcolsep}{0.1cm}
\centering
\scriptsize

\scalebox{1.0}{
\begin{tabular}{p{1.0cm}p{1.3cm}<{\centering}|@{}p{0.65cm}<{\centering}@{}p{0.6cm}<{\centering}@{}p{0.50cm}<{\centering}|@{}p{0.65cm}<{\centering}@{}p{0.6cm}<{\centering}@{}p{0.50cm}<{\centering}|@{}p{0.65cm}<{\centering}@{}p{0.6cm}<{\centering}@{}p{0.50cm}<{\centering}|@{}p{0.65cm}<{\centering}@{}p{0.6cm}<{\centering}@{}p{0.50cm}<{\centering}|@{}p{0.65cm}<{\centering}@{}p{0.6cm}<{\centering}@{}p{0.50cm}<{\centering}|@{}p{0.65cm}<{\centering}@{}p{0.6cm}<{\centering}@{}p{0.50cm}<{\centering}}
			\toprule
			\multirow{3}{*}{}  & & \multicolumn{3}{c|}{\texttt{Cifar10}} & \multicolumn{3}{c|}{\texttt{GTSRB}} & \multicolumn{3}{c|}{\texttt{VGGFace2}} & \multicolumn{3}{c|}{\texttt{ImageNet10}} & \multicolumn{3}{c}{\texttt{ImageNet50}}& \multicolumn{3}{|c}{\texttt{ImageNet100}} \\
		\cmidrule{3-20}	
			 & & CA & BA & ASR & CA & BA & ASR & CA & BA & ASR & CA & BA & ASR & CA & BA & ASR & CA & BA & ASR \\
			\midrule
		\multicolumn{2}{c|}{Before Defense}    & 94.0 & 1.0 & 99.0 & 99.7 & 1.1 & 98.9 & 96.7 & 0.0 & 100. & 88.6 & 9.4 & 89.4 & 84.4 & 0.7 & 99.2 & 81.7 & 0.3 & 99.6\\
		\midrule
		 \multirow{2}{*}{DiffPure} & DDPM   & 74.5 & 64.5 & 16.2 & 74.2 & 41.0 & 44.8 & 51.9 & 32.9 & 34.4 & 71.1 & 66.5 & 5.5 & 57.4 & 52.7 & 0.9 & 51.8 & 53.9 & 0.6\\
           & SDE  & 75.5 & 63.9 & 15.6 & 71.7 & 42.8 & 44.4 & 52.8 & 33.6 & 35.5 & 73.2 & 67.6 & 5.1 & 54.1 & 57.2 & 1.1 & 52.8 & 53.8 & 0.7\\
             \midrule
		\multirow{2}{*}{Ours} & Base  & 92.9 & \hlt{90.3} & 0.9 & \hlt{99.7} & 95.5 & \hlt{0.7} & 93.7 & \hlt{92.6} & \hlt{0.9} & 79.6 & 79.3 & 3.0 & 60.7 & 68.8 & \hlt{0.5} & 57.4 & 70.4 & \hlt{0.4}\\
		&  Large   & \hlt{93.3} & \hlt{90.3} & \hlt{0.7} & \hlt{99.7} & \hlt{96.0} & 1.1 & \hlt{94.3} & 91.9 & 1.9 & \hlt{84.1} & \hlt{81.1} & \hlt{2.8} & \hlt{71.3} & \hlt{75.2} & 0.6 & \hlt{65.4} & \hlt{76.2} & \hlt{0.4}\\
			\bottomrule
		\end{tabular} }

\vspace{1mm}

\scalebox{1.0}{
\begin{tabular}{p{1.0cm}p{1.3cm}<{\centering}|p{0.71cm}<{\centering}p{0.74cm}<{\centering}p{0.71cm}<{\centering}|p{0.71cm}<{\centering}p{0.74cm}<{\centering}p{0.71cm}<{\centering}|p{0.71cm}<{\centering}p{0.74cm}<{\centering}p{0.71cm}<{\centering}|p{0.71cm}<{\centering}p{0.74cm}<{\centering}p{0.71cm}<{\centering}}
			\toprule
			\multirow{3}{*}{}  & & \multicolumn{3}{c|}{\texttt{Cifar10}-IAB} & \multicolumn{3}{c|}{\texttt{GTSRB}-IAB} & \multicolumn{3}{c|}{\texttt{VGGFace2}-Blended} & \multicolumn{3}{c}{\texttt{ImageNet10}-Blended} \\
		\cmidrule{3-14}	
			 & & CA & BA & ASR & CA & BA & ASR & CA & BA & ASR & CA & BA & ASR \\
			\midrule
		\multicolumn{2}{c|}{Before Defense}  &  94.0 & 2.0 & 98.0 & 98.2 & 1.4 & 96.2 &	2.4 & 97.6
 & 98.6 &  89.6 & 29.6 &	69.0 \\
		\midrule
		 \multirow{2}{*}{DiffPure} & DDPM   &  77.2	& 11.2 & 87.2 & 74.0 & 8.6 & 90.6 & 49.8 &	52.4 &	1.4  & 70.2	& 70.6	& \hlt{3.4}\\
           & SDE  &  79.2 & 14.0	& 84.4 & 74.0 & 7.6 &	91.8  & 54.6 & 52.2 &	\hlt{0.6}  & 73.4	& 70.6 &	3.6 \\
             \midrule
		\multirow{2}{*}{Ours} & Base & 93.2 & 77.0 &	16.2 & 97.8 &	\hlt{80.0} & \hlt{18.2} & 92.4	& \hlt{92.0} & 1.0 & 71.4 &	65.2 &	18.2 \\
		&  Large   & \hlt{93.6} &	\hlt{79.6} & \hlt{14.6} & \hlt{98.0} &	79.6 &	19.0 & \hlt{93.4}	& 91.0	& 1.6 & \hlt{81.8}	& \hlt{74.2}	& 11.4 \\
			\bottomrule
		\end{tabular} }
\vspace{0mm}

\vspace{0mm}
\end{table*}

\subsection{Backdoor Attack Settings}
We use BadNet attack~\cite{gu2019badnets} with different triggers, Label-Consistent backdoor attack (LC)~\cite{turner2019lc}, Input-Aware dynamic Backdoor attack (IAB)~\cite{nguyen2020IAB} and Blended attack~\cite{chen2017targeted} to build backdoored models. 

The triggers of BadNet attack are chosen from 1$\times1 \sim 3\times$3 white or color patches. In addition, we use several 15$\times$15 icons as triggers for \texttt{VGGFace2} and \texttt{ImageNet}. These commonly seen object icons and social media icons are more natural in the real-world application. The triggers of LC attack are 3$\times$3 checkerboards in the four images corners.  The triggers of Blended attack is 15$\times$15 random pixels. The triggers of IAB attack are random color curves or shapes. They are sample-specific, in that each image has its unique trigger pattern.

We randomly select 10\% training data to create backdoored images. Then we train a model until it has a sufficiently high accuracy on clean images and attack success rate on backdoored images. The backbone network for \texttt{Cifar10} and \texttt{GTSRB} is ResNet18~\cite{he2016resnet} from random initialization. The backbone network for \texttt{VGGFace2} and \texttt{ImageNet} is pretrained ResNet50~\cite{he2016resnet}. We also considered other backbone networks in Sec.~\ref{sec:arch}. For each setting of \texttt{Cifar10} and \texttt{GTSRB}, we report average results over 14 repeated experiments from different target labels or initializations. For the large \texttt{VGGFace2} and \texttt{ImageNet}, we reduce it to 6 repeat experiments.

\begin{table*}[!t]

\caption{Comparison with other image purification methods. ($^\diamond$: white-box; others: black-box.)}\label{tab:agg_results} 
\vspace{0mm}

\renewcommand{\tabcolsep}{0.1cm}
\centering
\scriptsize

\scalebox{1.0}{
\begin{tabular}{p{1.0cm}p{1.3cm}<{\centering}|@{}p{0.65cm}<{\centering}@{}p{0.6cm}<{\centering}@{}p{0.50cm}<{\centering}|@{}p{0.65cm}<{\centering}@{}p{0.6cm}<{\centering}@{}p{0.50cm}<{\centering}|@{}p{0.65cm}<{\centering}@{}p{0.6cm}<{\centering}@{}p{0.50cm}<{\centering}|@{}p{0.65cm}<{\centering}@{}p{0.6cm}<{\centering}@{}p{0.50cm}<{\centering}|@{}p{0.65cm}<{\centering}@{}p{0.6cm}<{\centering}@{}p{0.50cm}<{\centering}|@{}p{0.65cm}<{\centering}@{}p{0.6cm}<{\centering}@{}p{0.50cm}<{\centering}}
			\toprule
			\multirow{3}{*}{}  & & \multicolumn{3}{c|}{\texttt{Cifar10}} & \multicolumn{3}{c|}{\texttt{GTSRB}} & \multicolumn{3}{c|}{\texttt{VGGFace2}}  & \multicolumn{3}{c|}{\texttt{ImageNet10}}  & \multicolumn{3}{c|}{\texttt{ImageNet50}} & \multicolumn{3}{c}{\texttt{ImageNet100}}  \\
		\cmidrule{3-20}	
			 & & CA & BA & ASR & CA & BA & ASR & CA & BA & ASR & CA & BA & ASR & CA & BA & ASR & CA & BA & ASR\\
			\midrule
		\multicolumn{2}{c|}{Before Defense} & 93.3 & 0.9 & 99.0 & 98.5 & 1.4 & 98.6 & 95.5 & 0.0 & 100. & 89.5 & 9.7 & 89.2 & 84.0 & 0.5 & 99.4 & 82.3 & 0.2 & 99.8\\
		\midrule
		 \multirow{2}{*}{Februus$^\diamond$} & XGradCAM     & 91.6 & 87.0 & 7.0 & 65.4 & 50.9 & 38.0 & 65.5 & 89.5 & 5.8 & -- & -- &  -- & -- & -- & -- & -- & -- & -- \\
		  & GradCAM++    & 80.0 & 91.0 & 2.3 & 59.1 & 73.9 & 14.6 & 63.1 & 89.4 & 5.9 & -- & -- &  -- & -- & -- & -- & -- & -- & --  \\
		\midrule          
             \multirow{2}{*}{PatchCleanser} & Vanilla  & 89.9 & 43.9 & 55.0 & 95.0 & 10.0 & 89.7 & 93.0 & 43.0 & 56.9 & 84.5 & 58.0 & 37.1 & 79.6 & 45.7 & 49.4 & 78.9 & 43.4 & 52.3\\
             & Variant   & 57.6 & 86.1 & 1.9 & 13.3 & 80.8 & 1.5 & 50.7 & \hlt{94.7} & \hlt{0.0} & 62.0 & 80.8 & 4.1 & 54.1 & \hlt{79.3} & \hlt{0.4} & 52.1 & \hlt{78.0} & \hlt{0.1}\\
             \cline{1-2}
            \multirow{2}{*}{Blur} & Weak  & 91.5 & 14.0 & 84.9 & \hlt{98.4} & 3.9 & 96.0 & \hlt{95.5} & 0.1 & 100. & \hlt{88.4} & 14.4 & 83.9 & \hlt{83.3} & 4.9 & 94.3 & \hlt{81.2} & 3.2 & 96.1\\
             & Strong   & 63.6 & 60.0 & 6.4 & 97.7 & 94.9 & 1.8 & 95.2 & 10.4 & 89.4 & 84.8 & 34.2 & 60.9 & 79.2 & 49.1 & 39.3 & 76.0 & 51.6 & 33.1\\
            \cline{1-2}
            \multirow{2}{*}{ShrinkPad} & Weak   & 90.7 & 50.3 & 45.0 & 97.5 & 33.3 & 65.0 & 93.8 & 35.5 & 62.5 & \hlt{88.4} & 43.0 & 52.1 & 82.0 & 39.7 & 51.1 & 80.0 & 42.3 & 46.0\\
            & Strong   & 86.7 & 36.7 & 57.9 & 92.8 & 23.5 & 72.3 & 88.3 & 54.4 & 38.3 & 86.7 & 56.7 & 36.0 & 79.4 & 55.1 & 29.8 & 77.2 & 58.6 & 22.5\\
             \midrule
		\multirow{2}{*}{Ours} & Base   & 92.5 & 90.8 & 0.9 & 98.2 & 95.3 & \hlt{0.9} & 91.3 & 92.0 & 1.6 & 79.9 & 81.1 & 4.8 & 61.7 & 70.1 & 0.8 & 59.0 & 67.9 & 0.4\\
		&  Large   & \hlt{92.7} & \hlt{91.1} & \hlt{0.8} & \hlt{98.4} & \hlt{96.0} & \hlt{0.9} & 92.9 & 91.8 & 2.2 & 83.9 & \hlt{83.7} & \hlt{3.9} & 72.6 & 76.1 & 0.6 & 69.5 & 73.9 & 0.3\\
			\bottomrule
		\end{tabular} }

\vspace{2mm}

\end{table*}

\begin{table*}[!t]

  \caption{Comparison results on three challenging attacks. (\texttt{VF2} short for \texttt{VGGFace2}, and \texttt{IN10} short for \texttt{ImageNet10}.)}
\label{tab:IAB-LC-Blended}
\vspace{0mm}

\begin{center}
\renewcommand{\tabcolsep}{0.1cm}
\centering
\scriptsize

\scalebox{1.0}{
\begin{tabular}{p{1.0cm}p{1.3cm}<{\centering}|@{}p{0.65cm}<{\centering}@{}p{0.6cm}<{\centering}@{}p{0.50cm}<{\centering}|@{}p{0.65cm}<{\centering}@{}p{0.6cm}<{\centering}@{}p{0.50cm}<{\centering}|@{}p{0.65cm}<{\centering}@{}p{0.6cm}<{\centering}@{}p{0.50cm}<{\centering}|@{}p{0.65cm}<{\centering}@{}p{0.6cm}<{\centering}@{}p{0.50cm}<{\centering}|@{}p{0.65cm}<{\centering}@{}p{0.6cm}<{\centering}@{}p{0.50cm}<{\centering}|@{}p{0.65cm}<{\centering}@{}p{0.6cm}<{\centering}@{}p{0.50cm}<{\centering}}
\toprule
 \multirow{3}{*}{} & & \multicolumn{3}{c|}{\texttt{Cifar10}--IAB} & \multicolumn{3}{c|}{\texttt{GTSRB}--IAB} & \multicolumn{3}{c|}{\texttt{Cifar10}--LC} & \multicolumn{3}{c|}{\texttt{GTSRB}--LC} & \multicolumn{3}{c|}{\texttt{VF2}--Blended}  & \multicolumn{3}{c}{\texttt{IN10}--Blended}  \\
		  \cmidrule{3-20}
			& & CA & BA & ASR & CA & BA & ASR & CA & BA & ASR & CA & BA & ASR & CA & BA & ASR & CA & BA & ASR \\
			\midrule
		\multicolumn{2}{c|}{Before Defense}  
        & 93.4 & 1.6 & 98.4 & 98.0 & 1.2 & 98.7 & 94.5 & 0.5 & 99.5 & 95.8 & 5.3 & 94.7 & 95.1 & 1.9 & 98.1 & 86.5 & 28.4 & 68.4\\
		\midrule		
		  \multirow{2}{*}{Februus$^\diamond$} & XGradCAM  & 91.7 & 29.9 & 68.1 & 68.4 & 72.5 & 24.8 & 92.6 & 63.7 & 33.9 & 80.6 & 91.7 & 5.1 & 68.9 & 74.6 & 21.4 & -- & -- & --\\
		 & GradCAM++   & 77.9 & 55.8 & 35.9 & 49.8 & \hlt{84.1} & 12.2 & 83.3 & 85.7 & 10.4 & 72.5 & 91.7 & 5.1 & 66.4 & 72.5 & 23.6 & -- & -- & --\\
        \midrule
        \multirow{2}{*}{PatchCleanser} & Vanilla  & 88.6 & 25.6 & 73.8 & 84.2 & 13.9 & 83.0 & 90.3 & 0.4 & 99.6 & 87.8 & 0.1 & 99.9 & 92.8 & 41.7 & 58.2 & 79.3 & 56.1 & 38.2\\
        & Variant  & 62.2 & 66.7 & 26.3 & 16.6 & 83.0 & \hlt{6.7} & 56.5 & 4.7 & 95.3 & 9.8 & 6.0 & 94.0 & 47.0 & 93.4 & 1.4 & 56.5 & 68.7 & 15.5\\
        \midrule
         \multirow{2}{*}{Blur} & Weak   & 91.3 & 33.9 & 63.7 & 97.8 & 18.6 & 81.2 & 92.5 & 92.3 & 0.7 & \hlt{95.5} & \hlt{95.1} & 1.2 & \hlt{95.1} & 38.0 & 60.7 & \hlt{86.5} & 47.5 & 46.8\\
        & Strong & 63.0 & 53.1 & 8.4 & 96.8 & 47.7 & 50.7 & 56.8 & 56.1 & 2.5 & 93.7 & 93.6 & \hlt{0.7} & 94.9 & \hlt{94.7} & \hlt{0.3} & 82.6 & 73.6 & 12.8\\
  	\midrule
        \multirow{2}{*}{ShrinkPad} & Weak   & 91.2 & 64.5 & 28.8 & 97.1 & 43.7 & 55.1 & 92.4 & 1.4 & 98.6 & 93.6 & 5.9 & 94.1 & 93.5 & 88.6 & 5.1 & 86.3 & \hlt{83.3} & \hlt{4.8}\\
        & Strong   & 88.5 & 80.6 & \hlt{7.1} & 93.1 & 62.2 & 32.5 & 89.7 & 85.1 & 6.1 & 81.8 & 68.0 & 23.1 & 87.0 & 86.1 & 1.0 & 84.5 & 80.3 & 5.4\\
  	\midrule
	   \multirow{2}{*}{Ours} & Base  & 93.0 & \hlt{81.8} & 10.5 & 97.8 & 76.2 & 21.4 & 93.7 & 94.1 & \hlt{0.4} & 93.9 & 93.6 & 2.3 & 90.7 & 91.8 & 1.0 & 73.4 & 68.0 & 17.3\\
	   & Large   & \hlt{93.1} & 80.0 & 13.0 & \hlt{98.0} & 70.6 & 27.4 & \hlt{93.9} & \hlt{94.3} & \hlt{0.4} & 94.8 & 93.5 & 2.4 & 92.1 & 92.2 & 0.9 & 81.2 & 73.1 & 14.3\\
	\bottomrule
		\end{tabular} }
	\end{center}	
 \vspace{0mm}

\end{table*}

\subsection{Method Configurations}
We avoid tuning our method to some specific dataset or attack. Instead, we use the \textbf{same} set of hyper-parameters for all experiments. The motivation is that as only one test image is available in the task, it is unlikely to tune those hyper-parameters reliably. Specifically, the default masking ratio is 75\%. $N_o=N_i=5$ and $N_r=10$. The initial thresholds used in our work is $\{0.6,0.55,0.5,0.45,0.4\}$. Even though, it is worthwhile mentioning that the image resolution and backdoor trigger patches can be highly diverse, better performances of our methods are expected with better tuned hyper-parameters. 

We use two pretrained Masked Autoencoders~\cite{he2022masked} that are available from their official repository. The \texttt{Base} variant has 12 encoder layers, and the \texttt{Large} variant has 24 encoder layers with an increased hidden size dimension. For \texttt{Cifar10} and \texttt{GTSRB}, we up-sample each image to 224$\times$224 first in order to fit MAE models. Afterwards, the MAE restorations are down-sampled back to the original image size.

\subsection{Experiment Environment}
We experiment with Nvidia A5000 or A6000 GPUs using PyTorch 1.8. For the implementation, the MAE related code is adapted from its official repository.

\begin{figure*}[!t]
	\centering	
	\begin{subfigure}{0.33\textwidth}
	\includegraphics[width=\linewidth]{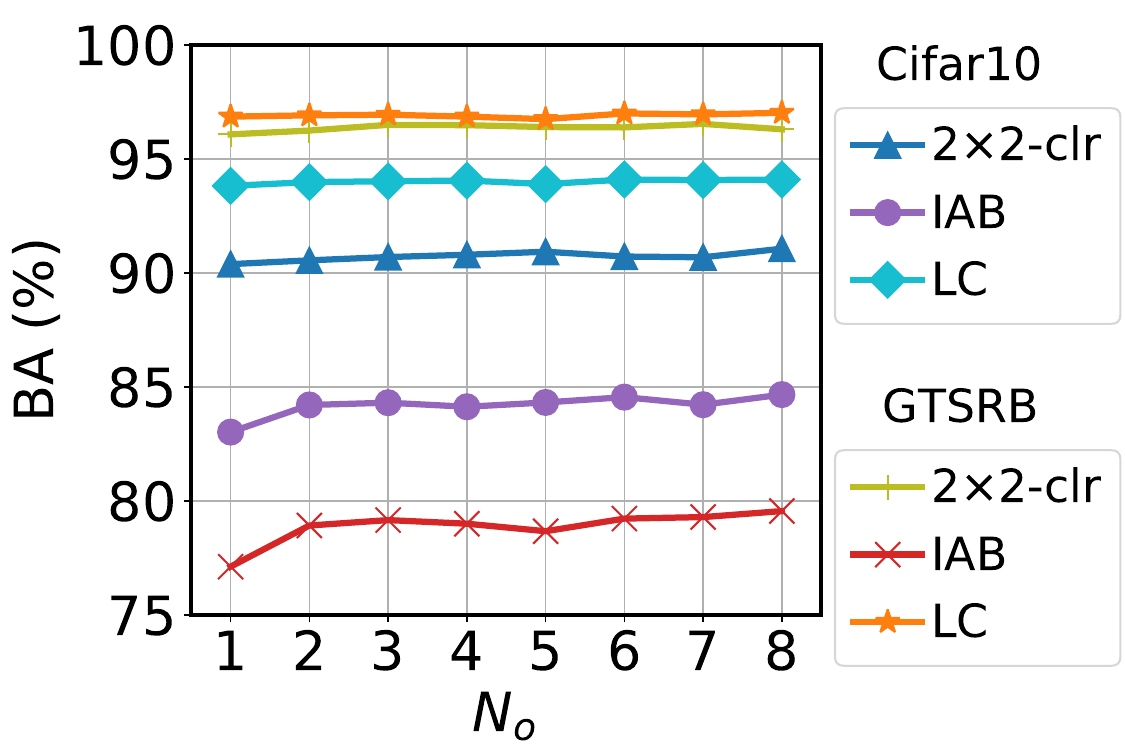}
	\caption{\texttt{Base}-$i$}
	\label{fig:abl:i}
        \end{subfigure}   
	\begin{subfigure}{0.31\textwidth}
	\includegraphics[width=\linewidth]{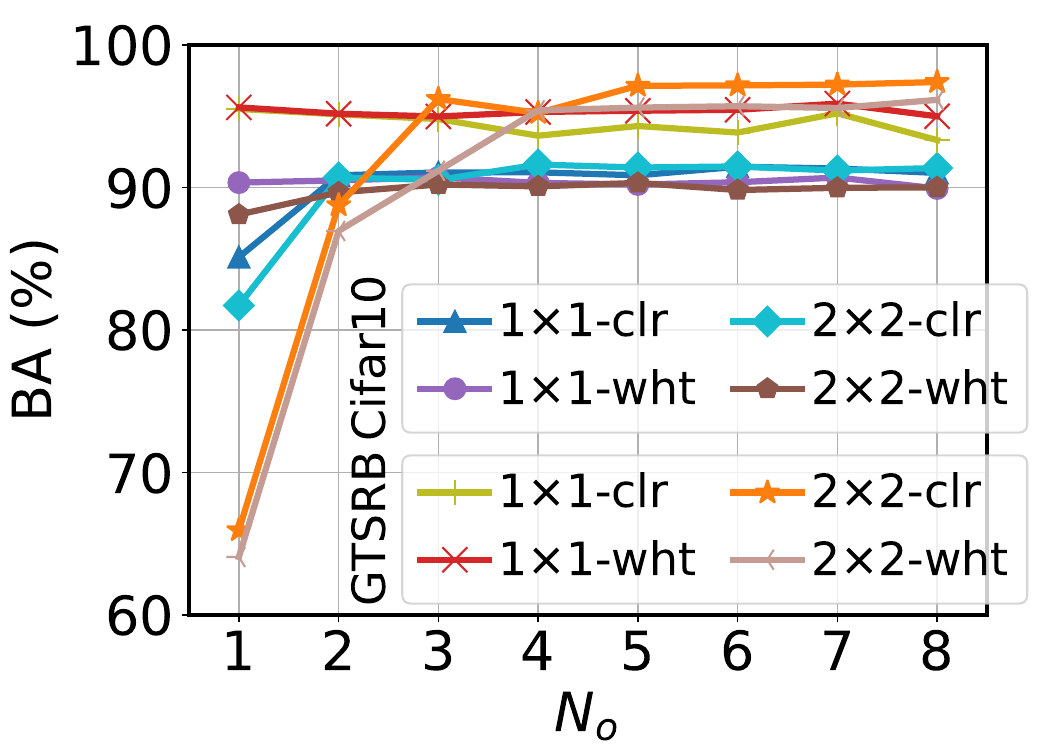}
	\caption{\texttt{Base}-$l$}
	\label{fig:abl:l}
	\end{subfigure}
 \begin{subfigure}{0.31\textwidth}
  \includegraphics[width=\linewidth]{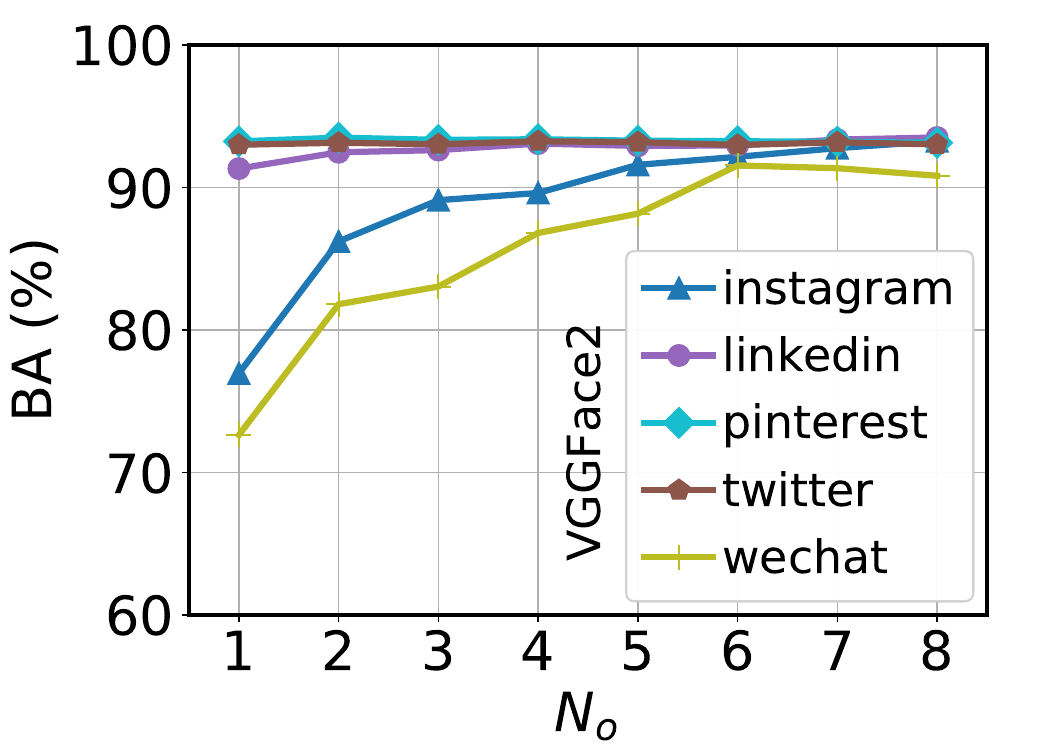}
  \caption{Full method}
	\label{fig:abl:full}
	\end{subfigure}

\begin{subfigure}{0.305\textwidth}
	\includegraphics[width=\linewidth]{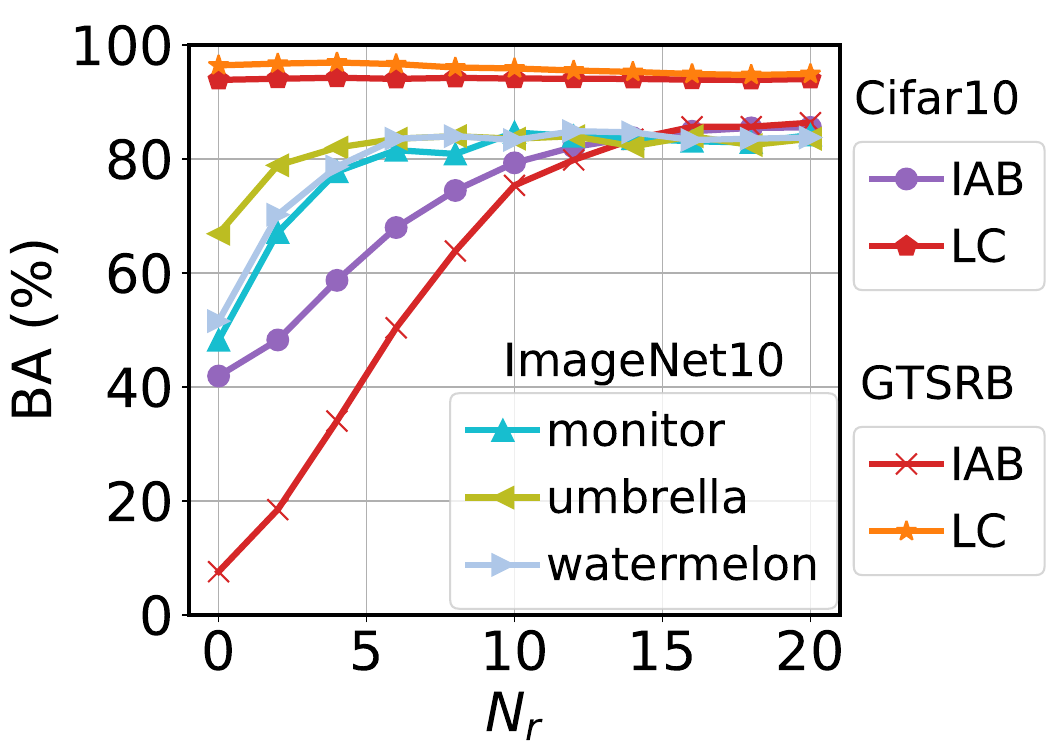}
	\caption{Full method}
	\label{fig:abl:rf}
    \end{subfigure}	
     \begin{subfigure}{0.305\textwidth}
 \includegraphics[width=\linewidth]{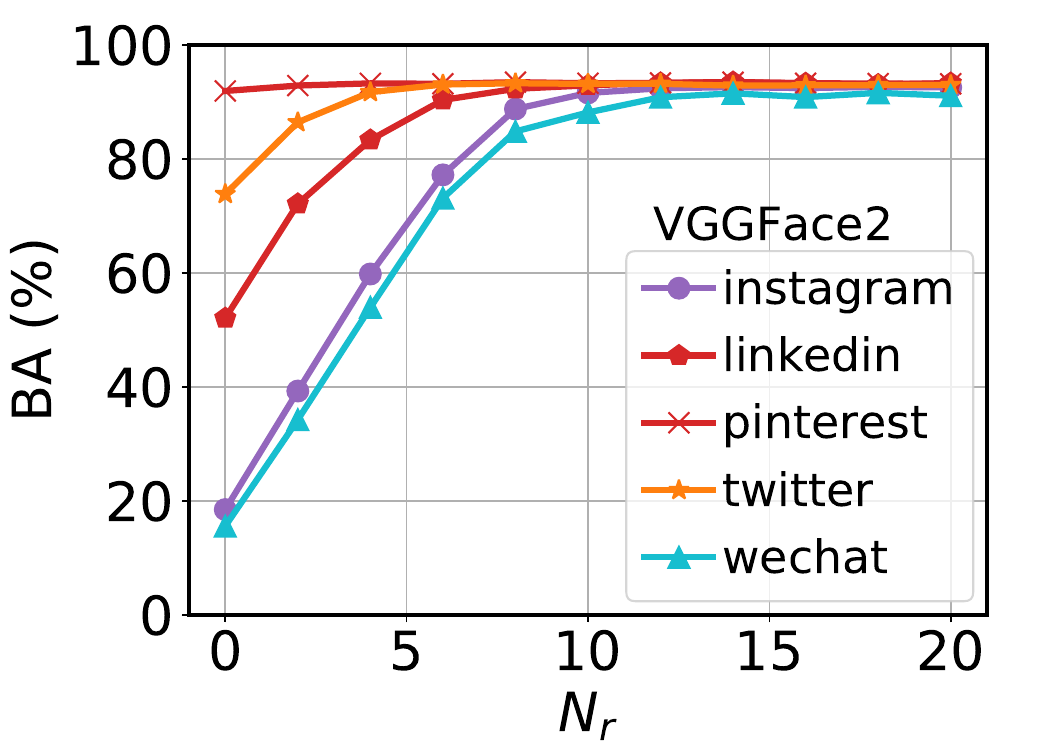}
 \caption{Full method}
 \label{fig:abl:rf_vgg}
 	\end{subfigure}	
 \vspace{0mm}
     \begin{subfigure}{0.365\textwidth}
 \includegraphics[width=\linewidth]{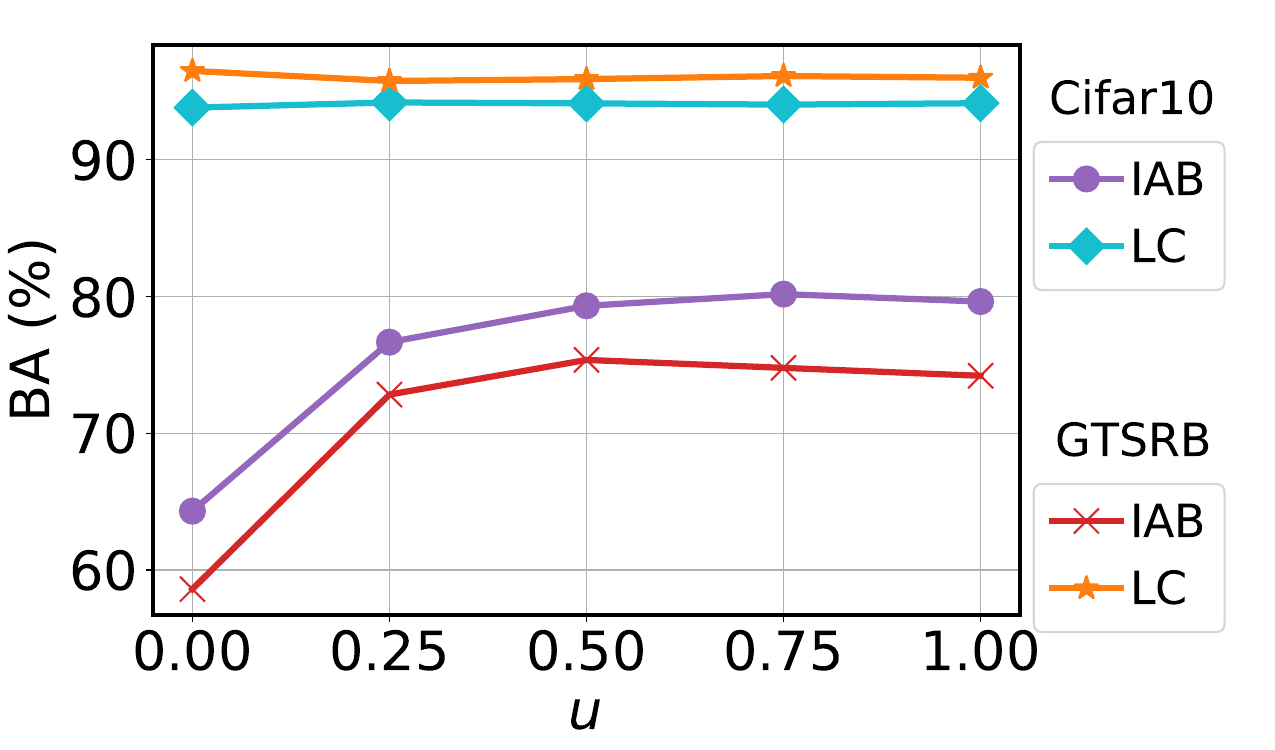}
\caption{Full method}
 \label{fig:abl:topo}
 	\end{subfigure}	
\caption{Effects of repeated times $N_o$, refinement times $N_r$, and sampling parameter $u$ in topology-aware token mask generation.}
\label{fig:abl}
\vspace{0mm}
\end{figure*}

\vspace{0mm}\cparagraph{Baseline methods.} We compare with several different lines of methods. \textbf{Blur} and \textbf{ShrinkPad}~\cite{li2021backdoor} are two simple purification methods based on test-time image transformation. More transformations are discussed in Sec.~\ref{sec:ttt}. \textbf{PatchCleanser}~\cite{xiang2022patchcleanser} is a certifiably robust defense method against adversarial patches via double-masking. \textbf{DiffPure}~\cite{nie2022diffusion} uses diffusion models for adversarial purification. In addition to these black-box methods, we also compare with a white-box method \textbf{Februus}~\cite{doan2020februus} that uses GradCAM~\cite{selvaraju2017grad} to locate triggers and restores missing parts with GAN models~\cite{goodfellow2014generative}.


\vspace{0mm}\cparagraph{Evaluation metrics} include the classification accuracy on clean images (CA) and backdoored images (BA), attack success rate (ASR). Due to page limit, we report results averaged over all triggers, and leave detailed results in Suppl.

\subsection{Main Results}
\vspace{0mm}\textbf{Comparison with diffusion model based DiffPure.}  Since the diffusion sampling process is extremely slow, we only report results on 500 test images in Tab.~\ref{tab:dp_results}. Overall, DiffPure can partially purify backdoored images but is much less effective than ours. DDPM and SDE sampling strategies obtain comparable performances. The low CA of DiffPure may be due to its reverse generative process that alternates image content, \eg, on \texttt{VGGFace2} where face recognition heavily relies on fine-grained attributes. Another observation is high ASR of DiffPure. This method is originally proposed for imperceptible adversarial perturbation, and the backdoor triggers are hard to be completely removed with diffusion sampling. More details and analysis are provided in Sec.~\ref{sec:visualization_results}.

\vspace{0mm}\cparagraph{Comparison with other purification methods.} 
Table~\ref{tab:agg_results} lists results of other methods. For Februus, we substitute its original GradCAM with two recent improvements to work on complex backbone networks. The GAN models are released by the authors, yet unavailable for \texttt{ImageNet}. Februus successfully purifies backdoored images but it is not black-box. Its performance is sensitive to CAM visualization. PatchCleanser uses two rounds of masking to locate the trigger patch. Its inconsistency check step is frequently affected by noisy label predictions, leading to low BA and high ASR. We propose a variant that can make decent predictions on backdoored images, but at a cost of much lower accuracies on clean images. The two simple test-time image transformations, Blur and ShrinkPad, both face a trade-off between CA and BA. Using a strong transformation is more likely to incapacitate backdoor triggers, but decreases clean accuracies. 

Our method achieves high accuracies on both clean and backdoored images. For the two variants, using MAE-Large performs slightly better due to better restorations. Unlike Blur and ShrinkPad that apply global transformations, our method first locates possible triggers and then restore the trigger region only. Compared with Februus and PatchCleanser, our method leverages MAE model to better locate triggers. These two points are key to our excellent performance. We also want to highlight that Tab.~\ref{tab:agg_results} reports the aggregated results. Using different sizes of backdoor triggers may lead to different observations of these methods. Please refer to Suppl. for more discussions.

\vspace{0mm}\cparagraph{Results on more challenging attacks.} 
In additional to the commonly used BadNet attack with different triggers, we consider three more challenging attacks. IAB attack triggers are sample-specific irregular color curves or shapes, often split into a few fragments. LC attack triggers are checkerboards at the four corners. Blended attack triggers are invisible noise patches in the lower right corner. From Tab.~\ref{tab:IAB-LC-Blended}, IAB and LC are more challenging for the comparison methods. The assumption of triggers covered by a small rectangle mask is invalid in PatchCleanser. The performances of comparison methods are rather inconsistent across different settings. Blur and ShrinkPad happen to be suitable choices for the invisible Blended attack. For all these challenging attack settings, our method obtains consistently high accuracies.

\section{Analysis}

\subsection{Effects of Topology-aware Refinement} The topology-aware refinement is vital to the generalizability of our method. It exploits initialized scores, and generates topology-aware token masks to refine the scores. This is beneficial especially to complex triggers. In Fig.~\ref{fig:topology}, the triggers are random curves and four distant checkerboards. Before refinement, the trigger regions have relatively high scores in $S^{(i)}$. But the contrast between trigger regions and clean regions are not significant. For each refinement, $\m_r$ is sampled in a topology-aware manner to be continuous patches. $S^{(i)}$ is updated to have increased values for tokens masked by $\m_r$ and reduced values for the rest. After 10 refinements, $S^{(i)}$ well reflects the trigger regions. It is worth noting that the refinement focuses on the triggers related to backdoor behaviors. Even though the blue line remains in the purified `dog' image, the red line has been removed, thus it makes correct label prediction.

\begin{figure}[!t]
	\centering	
\includegraphics[width=0.4\textwidth]{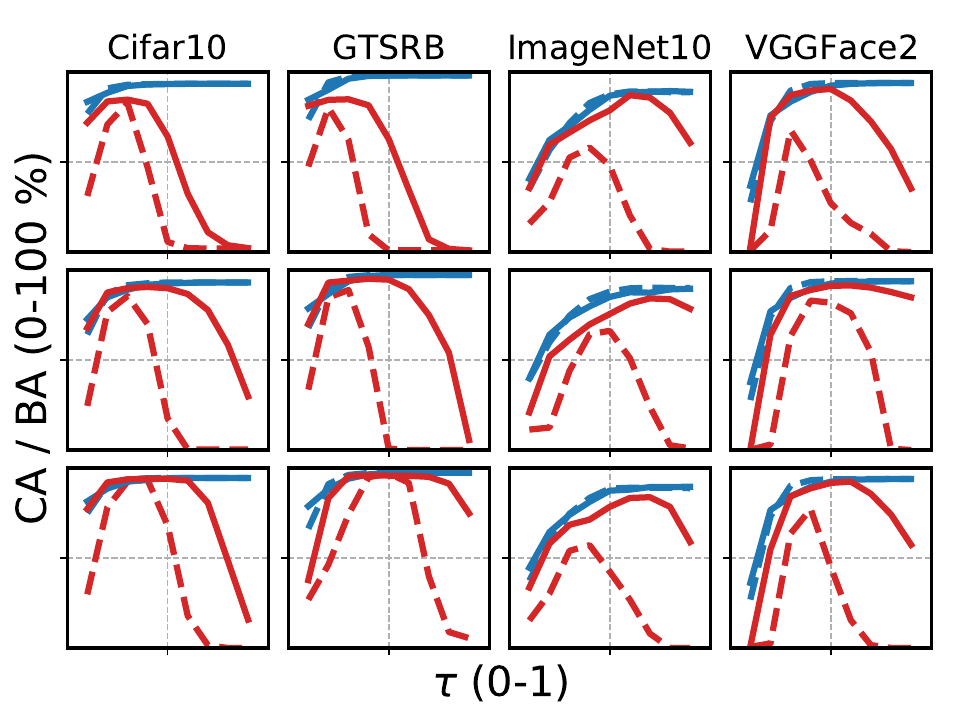}
\caption{Accuracies with fixed thresholds on backdoored (red) and clean (blue) images, before (dashed) or after (solid) refinement. Refinement enlarges ranges of optimal thresholds.}
\label{fig:abl:topology}
\end{figure}

In Fig.~\ref{fig:abl}, we find that $N_r=10$ is good enough for different triggers. One purpose of refinement is to increase contrast between scores of trigger regions and clean regions, so that the optimal threshold is easier to choose. In Fig.~\ref{fig:abl:topology}, we randomly select three defense tasks for each dataset. Instead of fusing restorations from multiple thresholds, we choose a fixed threshold ranging from 0.1 to 0.9, and plot the accuracy curves. In each subplot, red/blue lines denote backdoored/clean images, dashed/solid lines denote before/after refinement. We can see that before refinement, the optimal thresholds have narrow ranges and vary across tasks. After refinement, they become wider. It is thus easy to set unified thresholds for different tasks.

\begin{table*}[!t]

\caption{Running time (in minutes) of comparison test-time defense methods using one A6000 GPU. ($^\dagger$Since DiffPure is extremely time-consuming, we randomly select 1k images for testing and estimate running time on 19k.)}

\renewcommand{\tabcolsep}{0.06cm}
\begin{center}
\scriptsize
 \scalebox{1.0}{
	\begin{tabular}{p{1.5cm}<{\centering}p{1.5cm}<{\centering}p{1.5cm}<{\centering}|p{1.5cm}<{\centering}p{1.5cm}<{\centering}p{1.5cm}<{\centering}p{1.5cm}<{\centering}p{1.5cm}<{\centering}p{1.5cm}<{\centering}p{1.5cm}<{\centering}}
		\toprule
		  Dataset & \# Test Images & Backbone & Februus & PatchCleanser & Blur & ShrinkPad & DiffPure & Ours-Base & Ours-Large \\
		\midrule
		\texttt{Cifar10} & 19k & ResNet-18 & 10.6 & 24.5 & 0.4 & 0.5 & $>$3000$^\dagger$ & 28.1 & 54.0\\		
		\bottomrule
	\end{tabular} }
    \end{center}

\label{tab:time}
\end{table*}

\begin{figure*}[!t]
	\centering	
	\includegraphics[width=0.8\linewidth]{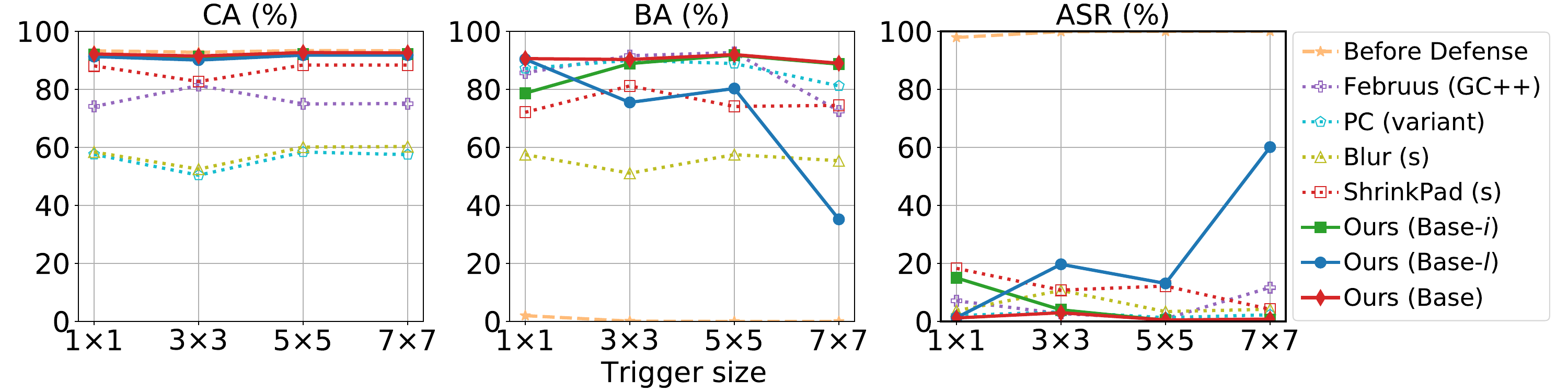}
	\caption{Varying trigger (checker-board) sizes on \texttt{Cifar10}.}\label{fig:cifar10_vs}	
\end{figure*}

\subsection{Sensitivity on Hyper-parameters} Our method mainly involves two critical hyper-parameters, the repeated times $N_o$ and the refinement times $N_r$. Throughout the experiments, we use $N_o=5$ and $N_r=10$. Figure~\ref{fig:abl} plots their effects. For the image-based score $S^{(i)}$, the SSIM score map is similar for different MAE restorations. Thus averaging over 2 repeated results is good enough. For the label-based score $S^{(l)}$, averaging over many repeated results reduces the variance. $N_o=5$ generally performs well for both scores. Using larger refinement times $N_r$ leads to higher accuracies. $N_r=10$ is good enough for different datasets and triggers.

\subsection{Topology-aware Token Mask Generation}\label{sec:topology}
In the score refinement, we generate topology-aware token masks. We repeatedly choose $t_i=\arg\max_{t_k} (S^{*}[t_k]+u\llbracket t_k \in \texttt{Adj}(\mathcal{T}) \rrbracket )\cdot \sigma_{k} $ with $u=0.5$, where $\texttt{Adj}(\mathcal{T})$ includes all 4-nearest neighbors of tokens in $\mathcal{T}$ and $\sigma_{k}\sim U(0,1)$ is a random variable. Here $u$ is the additional probability assigned to the neighboring tokens. When $u=0$, the sampling procedure only select tokens with highest trigger scores. To see the effect of $u$, Fig.~\ref{fig:abl:topo} plots the results on four defense tasks with increasing $u$. For the challenging IAB attacks, the performances drops when not using topology-aware sampling (\ie, $u=0$). $u=0.5$ obtains relatively good performances on the four tasks. Note that due to the existence of random variable $\sigma_k$, using $u=1.0$ still leads to some randomness in the token selection.

\captionsetup[subfigure]{labelformat=empty}
\begin{figure*}[!t]
	\centering
\begin{subfigure}{0.16\linewidth}
	\includegraphics[width=\textwidth]{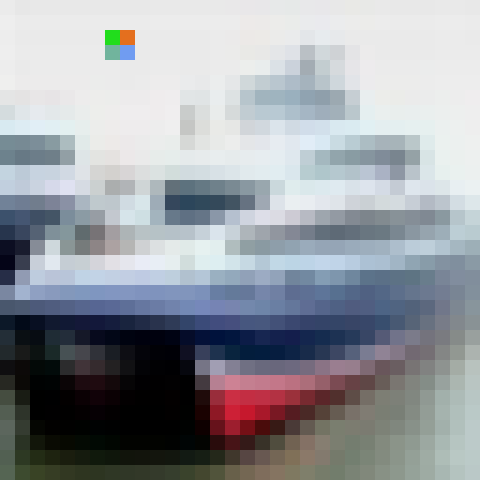}
\end{subfigure}
\begin{subfigure}{0.16\linewidth}
	\includegraphics[width=\textwidth]{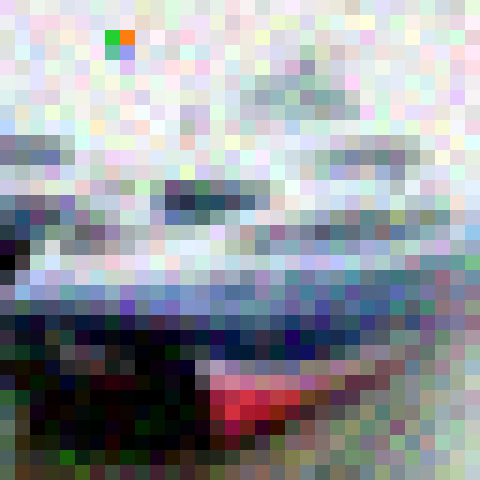}
\end{subfigure}
\begin{subfigure}{0.16\linewidth}
	\includegraphics[width=\textwidth]{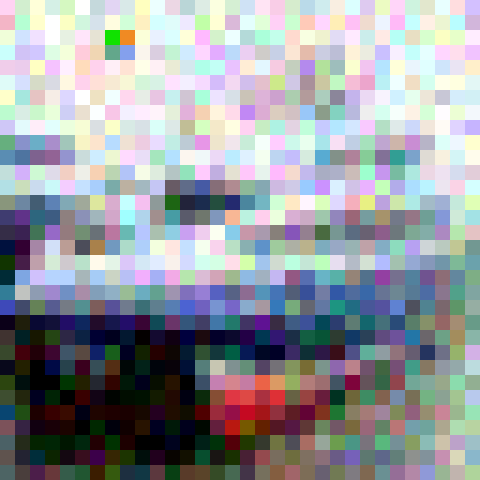}
\end{subfigure}
\begin{subfigure}{0.16\linewidth}
	\includegraphics[width=\textwidth]{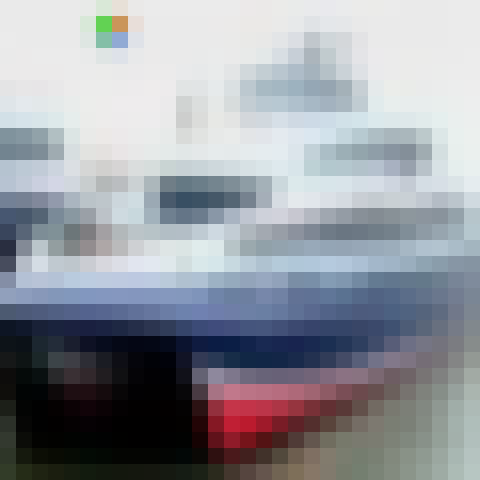}
\end{subfigure}
\begin{subfigure}{0.16\linewidth}
	\includegraphics[width=\textwidth]{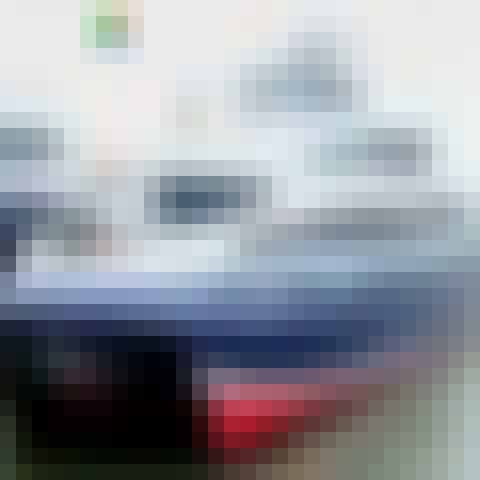}
\end{subfigure}
\begin{subfigure}{0.16\linewidth}
	\includegraphics[width=\textwidth]{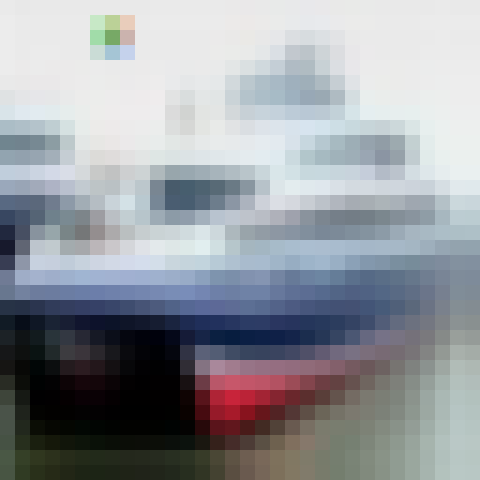}
\end{subfigure}
\begin{minipage}{0.16\linewidth}
        \scriptsize
	\centering Original Image \\ CA = 93.09 \% \\ BA = 0.00 \% \\ ASR = 100. \% 
\end{minipage} 
\begin{minipage}{0.16\linewidth}
        \scriptsize
	\centering Gaussian Noise (w) \\ CA = 53.57 \%  \\ BA = 0.09 \%  \\ ASR = 99.82 \% 
\end{minipage} 
\begin{minipage}{0.16\linewidth}
        \scriptsize
	\centering  Gaussian Noise (s)  \\ CA = 14.53 \%  \\ BA = 3.90 \%  \\ ASR = 73.10 \% 
\end{minipage} 
\begin{minipage}{0.16\linewidth}
        \scriptsize
	\centering  Gaussian Blur (w) \\ CA = 91.3 \%  \\ BA = 0.30 \%  \\ ASR = 99.68 \% 
\end{minipage} 
\begin{minipage}{0.16\linewidth}
        \scriptsize
	\centering  Gaussian Blur (s)  \\ CA = 63.15 \%  \\ BA = 61.98 \%  \\ ASR = 2.01 \% 
\end{minipage} 
\begin{minipage}{0.16\linewidth}
        \scriptsize
	\centering  Optical Distortion \\ CA = 86.78 \%  \\ BA = 69.94 \%  \\ ASR =  20.00 \% 
\end{minipage} 
\vspace{2mm}

\begin{subfigure}{0.16\linewidth}
	\includegraphics[width=\textwidth]{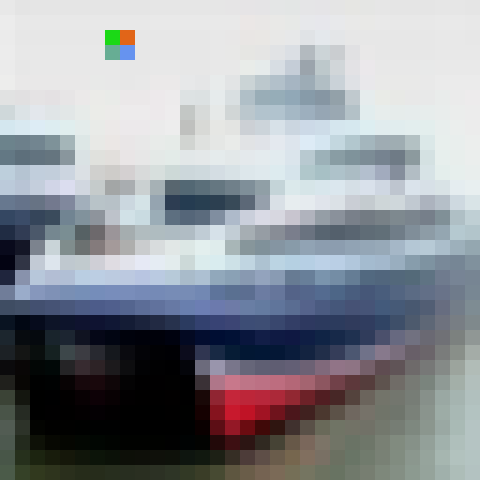}
\end{subfigure}
\begin{subfigure}{0.16\linewidth}
	\includegraphics[width=\textwidth]{fig-supp/cifar10_badnet_2by2_color_img1_RandomGamma_0.png}
\end{subfigure}
\begin{subfigure}{0.16\linewidth}
	\includegraphics[width=\textwidth]{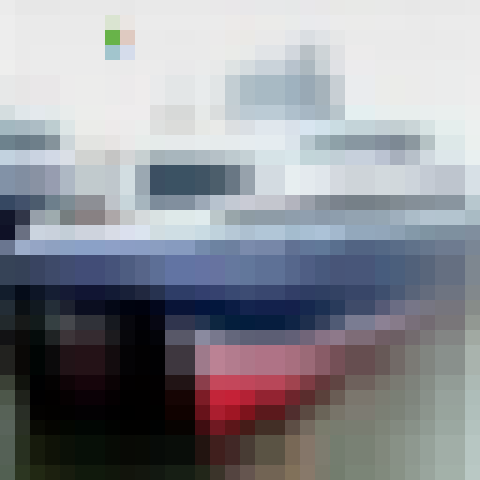}
\end{subfigure}
\begin{subfigure}{0.16\linewidth}
	\includegraphics[width=\textwidth]{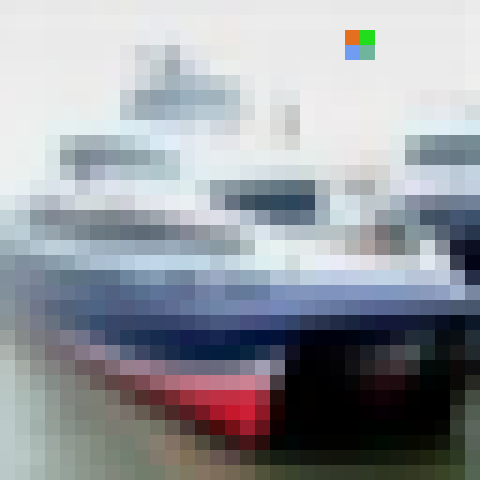}
\end{subfigure}
\begin{subfigure}{0.16\linewidth}
	\includegraphics[width=\textwidth]{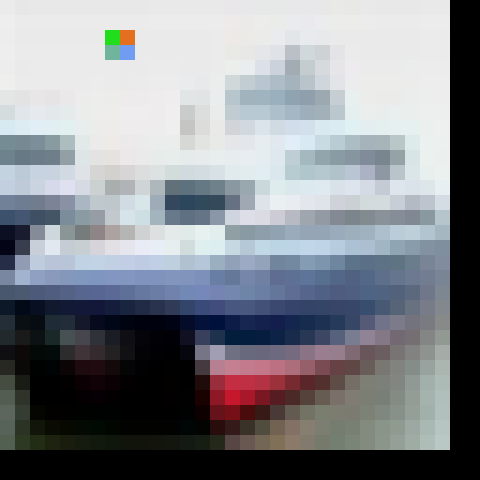}
\end{subfigure}
\begin{subfigure}{0.16\linewidth}
	\includegraphics[width=\textwidth]{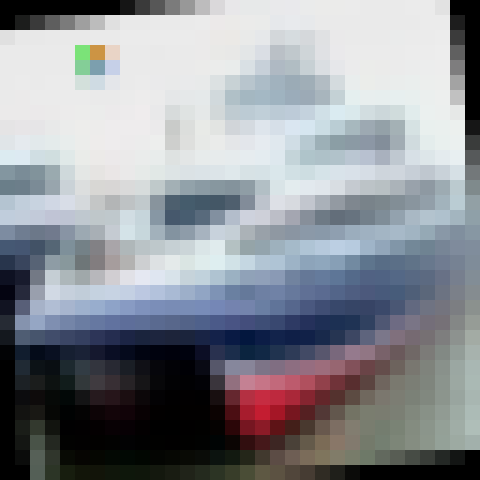}
\end{subfigure}
\begin{minipage}{0.16\linewidth}
        \scriptsize
	\centering Random Contrast \\ CA = 93.01 \%  \\ BA = 0.00 \%  \\ ASR = 100.  \% 
\end{minipage} 
\begin{minipage}{0.16\linewidth}
        \scriptsize
	\centering Random Gamma  \\ CA = 93.07 \%  \\ BA = 0.00 \%  \\ ASR =  100. \% 
\end{minipage} 
\begin{minipage}{0.16\linewidth}
        \scriptsize
	\centering  Grid Distortion  \\ CA = 73.76 \%  \\ BA = 26.74 \%  \\ ASR =  64.61 \% 
\end{minipage} 
\begin{minipage}{0.16\linewidth}
        \scriptsize
	\centering  Horizontal Flip \\ CA = 93.33 \%  \\ BA = 0.00 \%  \\ ASR = 100.  \% 
\end{minipage} 
\begin{minipage}{0.16\linewidth}
        \scriptsize
	\centering  Down Scale  \\ CA = 91.16 \%  \\ BA = 0.00 \%  \\ ASR =  100. \% 
\end{minipage} 
\begin{minipage}{0.16\linewidth}
        \scriptsize
	\centering  Affine Trans  \\ CA = 89.37 \%  \\ BA = 1.10 \%  \\ ASR = 98.79  \% 
\end{minipage}
\vspace{2mm}
\caption{Defense results of applying test-time image transformations on \texttt{Cifar10} with 2$\times$2-color trigger. The metrics shown are calculated on the entire test set.}
\label{fig:test_transformation}
\vspace{2mm}
\end{figure*}

\begin{table*}[!t]
\caption{Varying backbone network architectures.}
\begin{center}
\renewcommand{\tabcolsep}{0.1cm}
\centering
\scriptsize
\scalebox{0.95}{
\begin{tabular}{p{1.0cm}p{2.0cm}<{\centering}|p{0.5cm}<{\centering}p{0.5cm}<{\centering}p{0.5cm}<{\centering}|p{0.5cm}<{\centering}p{0.5cm}<{\centering}p{0.5cm}<{\centering}|p{0.5cm}<{\centering}p{0.5cm}<{\centering}p{0.5cm}<{\centering}|p{0.5cm}<{\centering}p{0.5cm}<{\centering}p{0.5cm}<{\centering}|p{0.5cm}<{\centering}p{0.5cm}<{\centering}p{0.5cm}<{\centering}|p{0.5cm}<{\centering}p{0.5cm}<{\centering}p{0.5cm}<{\centering}}
	\toprule
        & & \multicolumn{9}{c|}{\texttt{Cifar10}} & \multicolumn{9}{c}{\texttt{VGGFace2}} \\
        \cmidrule{3-20}	
	\multirow{3}{*}{} & & \multicolumn{3}{c|}{ResNet18} & \multicolumn{3}{c|}{6 Conv + 2 Dense} & \multicolumn{3}{c|}{VGG16}  & \multicolumn{3}{c|}{ResNet18} & \multicolumn{3}{c|}{ResNet50} & \multicolumn{3}{c}{VGG16} \\
	\cmidrule{3-20}	
			& & CA & BA & ASR & CA & BA & ASR & CA & BA & ASR & CA & BA & ASR & CA & BA & ASR & CA & BA & ASR \\
	\midrule
	\multicolumn{2}{c|}{Before Defense}  & 92.8 & 0.1 & 99.9 & 91.3 & 0.0 & 100. & 89.8 & 0.0 & 100. & 94.0 & 0.0 & 100. & 95.5 & 0.0 & 100. & 91.5 & 0.0 & 100.\\
	\midrule	
	\multirow{2}{*}{Februus} & XGradCAM    & 91.0 & 83.9 & 11.0 & 86.2 & 90.0 & 2.7 & 87.3 & 45.7 & 50.0 & 46.3 & 93.6 & 0.2 & 65.5 & 89.5 & 5.8 & 81.3 & 75.6 & 17.7\\
	& GradCAM++  & 88.0 & 88.4 & 6.1 & 91.3 & 91.2 & 1.5 & 76.2 & 77.5 & 15.1 & 43.5 & 92.8 & 1.1 & 63.1 & 89.4 & 5.9 & 80.5 & 77.1 & 16.1\\
	\midrule
        \multirow{2}{*}{PatchCleanser} & Vanilla   & 89.2 & 33.1 & 66.8 & 87.0 & 36.8 & 62.9 & 86.5 & 29.7 & 70.1 & 92.0 & 36.4 & 63.6 & 93.0 & 43.0 & 56.9 & 88.3 & 35.6 & 64.2\\
        & Variant & 50.4 & 90.0 & 3.3 & 52.6 & 88.2 & 2.8 & 47.8 & 86.6 & 2.8 & 45.0 & 93.1 & 0.0 & 50.7 & 94.7 & 0.0 & 42.8 & 90.3 & 0.1\\
        \midrule
         \multirow{2}{*}{Blur} & Weak   & 90.5 & 90.0 & 2.3 & 87.5 & 73.2 & 20.0 & 86.2 & 27.5 & 69.6 & 93.9 & 0.0 & 100. & 95.5 & 0.1 & 100. & 88.9 & 69.5 & 22.8\\
        & Strong  & 52.5 & 51.1 & 10.7 & 55.0 & 53.7 & 7.3 & 54.7 & 53.6 & 6.8 & 93.7 & 14.2 & 85.4 & 95.2 & 10.4 & 89.4 & 88.3 & 78.8 & 11.3\\
  	\midrule
        \multirow{2}{*}{ShrinkPad} & Weak  & 86.8 & 24.5 & 74.5 & 83.4 & 23.5 & 75.8 & 82.1 & 1.6 & 98.3 & 91.8 & 12.1 & 87.3 & 93.8 & 35.5 & 62.5 & 88.2 & 5.4 & 93.8\\
        & Strong  & 82.7 & 81.2 & 10.8 & 73.8 & 70.5 & 23.4 & 74.1 & 72.1 & 16.3 & 83.5 & 24.9 & 71.1 & 88.3 & 54.4 & 38.3 & 72.6 & 25.2 & 52.3\\
  	\midrule
	\multirow{2}{*}{Ours} & Base  & 91.5 & 90.3 & 3.0 & 89.9 & 90.0 & 1.3 & 88.5 & 87.8 & 2.5 & 87.9 & 88.1 & 4.2 & 91.3 & 92.0 & 1.6 & 83.7 & 84.5 & 5.2\\
	& Large  & 91.7 & 91.1 & 2.1 & 90.3 & 90.2 & 1.2 & 88.8 & 88.3 & 2.2 & 90.5 & 88.0 & 4.6 & 92.9 & 91.8 & 2.2 & 85.5 & 85.3 & 4.5\\
			\bottomrule
		\end{tabular} }
	\end{center}	
\vspace{0mm}
\label{tab:sup:arch}
\end{table*}

\subsection{Computational Efficiency} 

Running time is an important factor for test-time defense. We conduct analysis on \texttt{Cifar10} using one A6000 GPU. The total number of test images is 19k, including 10k clean images and 9k backdoored images. As listed in Tab.~\ref{tab:time}, Blur and ShrinkPad run fast as they only need to do image transformation. However, they could not obtain high accuracies as shown in the paper. Februus takes about 10.6 minutes. It uses GradCAM++ to locate triggers and inpaint images with GAN models. PatchCleanser takes about 24.5 minutes. It searches for triggers with two round masking. Ours with \texttt{Base} MAE consumes similar time as PatchCleanser. When using \texttt{Large} MAE, it takes about 54.0 minutes. Overall, our methods can purify images within a reasonable time. For DiffPure, since its diffusion process is extremely slow, we randomly select 1k images for testing. The running time on 19k is estimated to be over 3000 minutes. This shows that our method is far more efficient than diffusion model-based method. 

The major computational overhead comes from using MAE models to generate restored images. This can be conducted for multiple test images or one test image with multiple MAE masks in parallel. Another feasible way is to distillate MAEs into smaller models or quantify their weights (\textit{e.g.}, binary values) for faster inference.

\subsection{Varying Trigger Size}\label{sec:trigger_size}
The trigger size affects the difficulty to detect these triggers. In the BadNet work~\cite{gu2019badnets}, the authors use 3$\times$3-checkerboard as triggers. In Fig.~\ref{fig:cifar10_vs}, we present defense results on \texttt{Cifar10} using various sizes of checkerboard triggers. The performances of comparison methods are discouraging. Our method maintains high accuracies on clean images. Our \texttt{Base-$i$} is not working well on 1$\times$1 trigger because it is hard to detect such a small trigger using image similarity. \texttt{Base-$l$}, on the contrary, works well on this small trigger using label consistency. As trigger size becomes larger, the performance of \texttt{Base-$l$} drops because the trigger can not be removed completely through random masking. The full method \texttt{Base} combines the merits of both image similarity and label consistency, and works for all cases.

\subsection{Defense with More Test-Time Transformations}\label{sec:ttt}
To defense against backdoor attack, test-time transformations have been used in some previous works~\cite{gao2019strip,sarkar2020backdoor,qiu2021deepsweep}. Since they are training free and can be applied to our task, we briefly summarize these methods and remark on their limitation in our blind backdoor defense setting. \textbf{Supression}~\cite{sarkar2020backdoor} creates multiple fuzzed copies of backdoored images, and uses majority voting among fuzzed copies to recover label prediction. The fuzzed copies are obtained by adding random uniform noise or Gaussian noise to the original image. However, the intensity of noise is critical. Weak noise would not remove the backdoor behaviour, while strong noise may destroy the semantic content.  \textbf{DeepSeep}~\cite{qiu2021deepsweep} mitigates backdoor attacks using data augmentation. It first fine-tunes the infected model via clean samples with an image transformation policy, and then preprocesses inference samples with another image transformation policy. The image transformation functions include affine transformations, median filters, optical distortion, gamma compression, \etc. The fine-tuning stage requires additional clean samples, which are unavailable in our setting. \textbf{STRIP}~\cite{gao2019strip} superposes a test image with multiple other samples, and observes the entropy of predicted labels of these replicas. It aims to detect backdoored inputs, but could not locate the triggers nor recover the true label.

In Fig.~\ref{fig:test_transformation}, we try different test-time image transformations on \texttt{Cifar10} with 2$\times$2-color trigger. For each transformation, we calculate the CA on clean images, BA and ASR on backdoored images. As can be seen, some weak transformations, like Gaussian Noise (w), Gaussian Blur (w), Random Contrast/Gamma, Horizontal Flip and Down Scale, can not reduce ASR. While the rest strong transformations reduces ASR, they also compromise accuracies on clean images unacceptably. To maintain performance on clean images, the model needs to adapt to these image transformations, \eg, through fine-tuning like DeepSeep does. Such requirement is infeasible in the blind backdoor defense, especially for black-box models.

\begin{figure*}[!t]
	\centering

\begin{subfigure}{0.12\textwidth}
	\includegraphics[width=\textwidth]{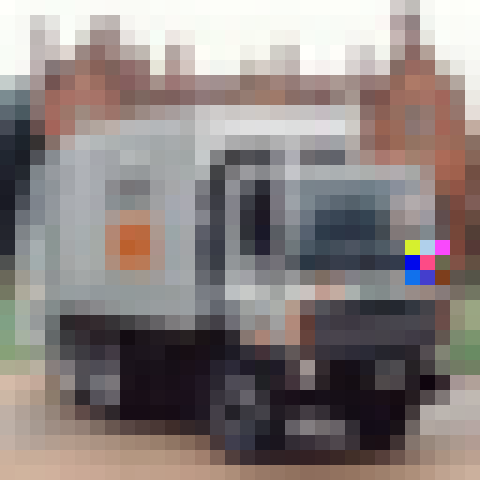}
\end{subfigure}
\begin{subfigure}{0.12\textwidth}
	\includegraphics[width=\textwidth]{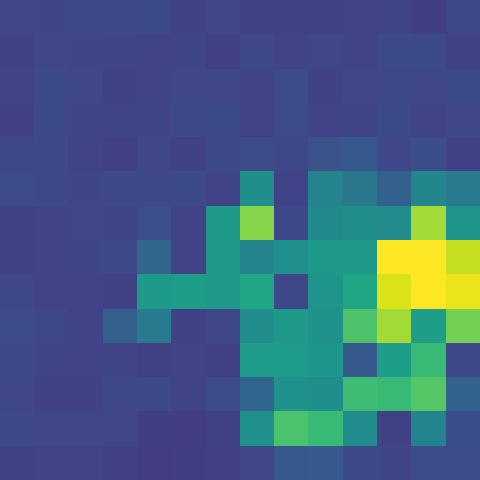}
\end{subfigure}
\begin{subfigure}{0.12\textwidth}
	\includegraphics[width=\textwidth]{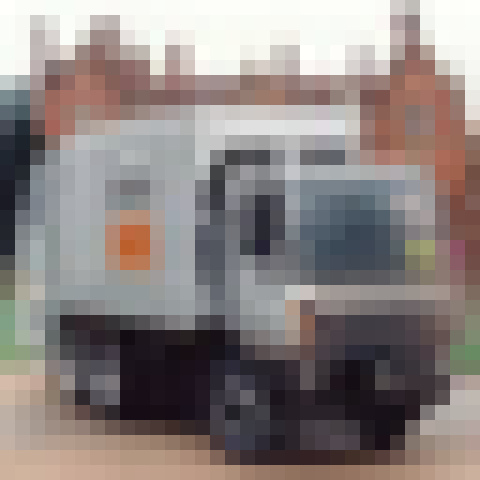}
\end{subfigure}
\begin{subfigure}{0.12\textwidth}
	\includegraphics[width=\textwidth]{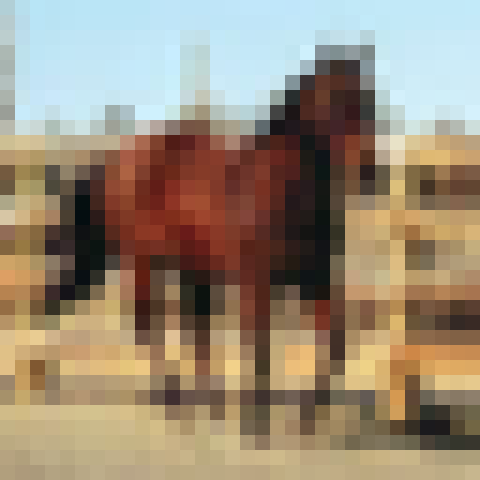}
\end{subfigure}
\begin{subfigure}{0.12\textwidth}
	\includegraphics[width=\textwidth]{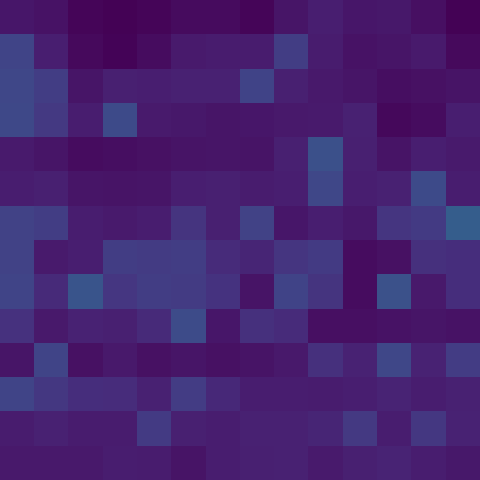}
\end{subfigure}
\begin{subfigure}{0.12\textwidth}
	\includegraphics[width=\textwidth]{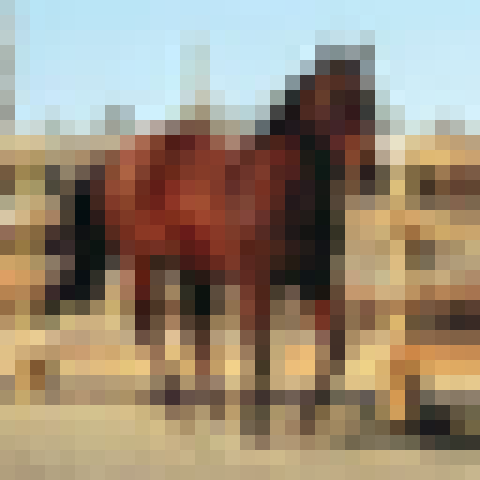}
\end{subfigure}
\VM

\begin{subfigure}{0.12\textwidth}
	\includegraphics[width=\textwidth]{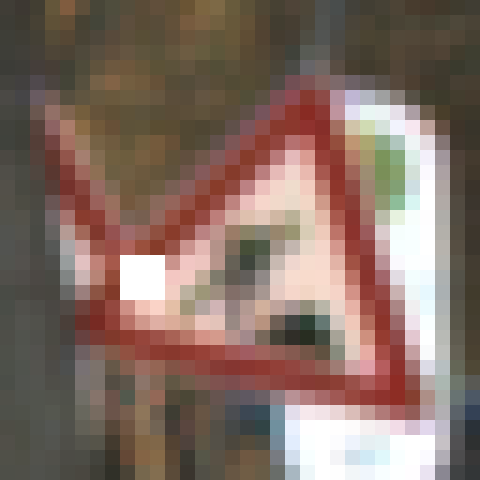}
\end{subfigure}
\begin{subfigure}{0.12\textwidth}
	\includegraphics[width=\textwidth]{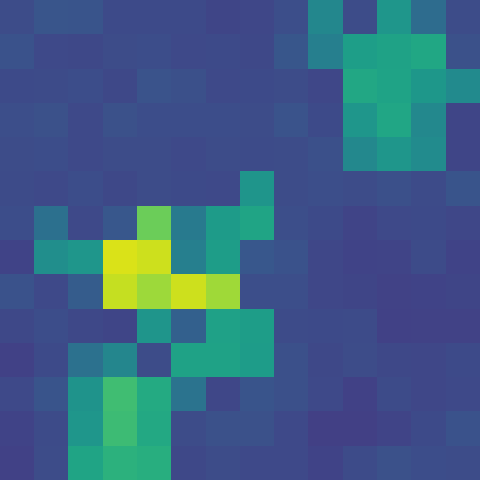}
\end{subfigure}
\begin{subfigure}{0.12\textwidth}
	\includegraphics[width=\textwidth]{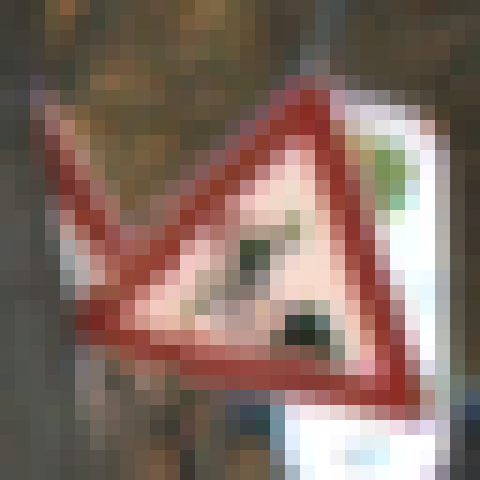}
\end{subfigure}
\begin{subfigure}{0.12\textwidth}
	\includegraphics[width=\textwidth]{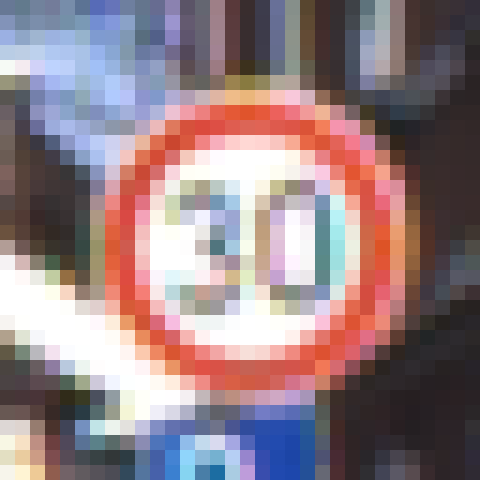}
\end{subfigure}
\begin{subfigure}{0.12\textwidth}
	\includegraphics[width=\textwidth]{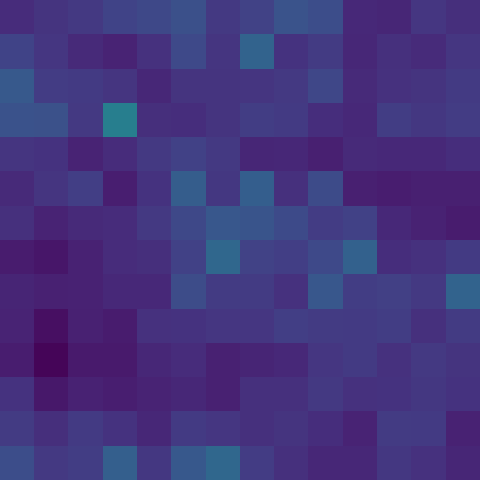}
\end{subfigure}
\begin{subfigure}{0.12\textwidth}
	\includegraphics[width=\textwidth]{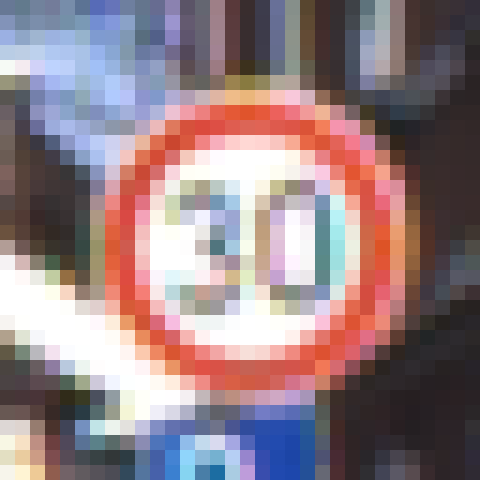}
\end{subfigure}
\VM

\begin{subfigure}{0.12\textwidth}
	\includegraphics[width=\textwidth]{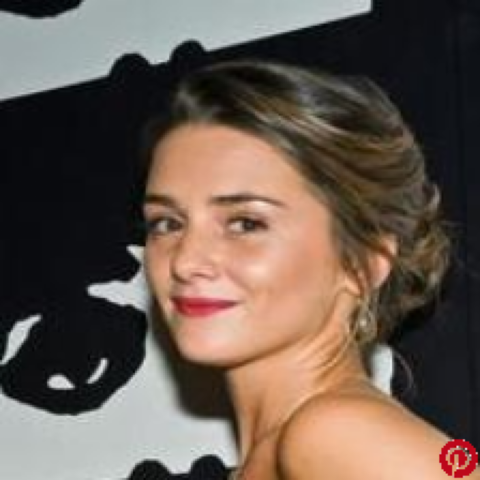}
\end{subfigure}
\begin{subfigure}{0.12\textwidth}
	\includegraphics[width=\textwidth]{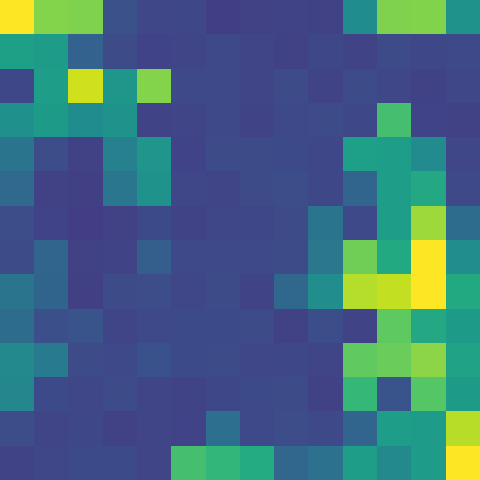}
\end{subfigure}
\begin{subfigure}{0.12\textwidth}
	\includegraphics[width=\textwidth]{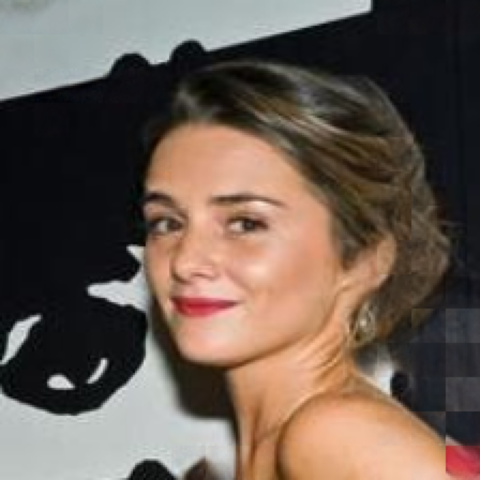}
\end{subfigure}
\begin{subfigure}{0.12\textwidth}
	\includegraphics[width=\textwidth]{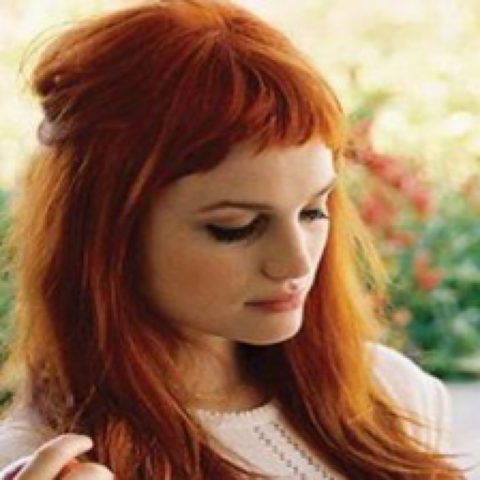}
\end{subfigure}
\begin{subfigure}{0.12\textwidth}
	\includegraphics[width=\textwidth]{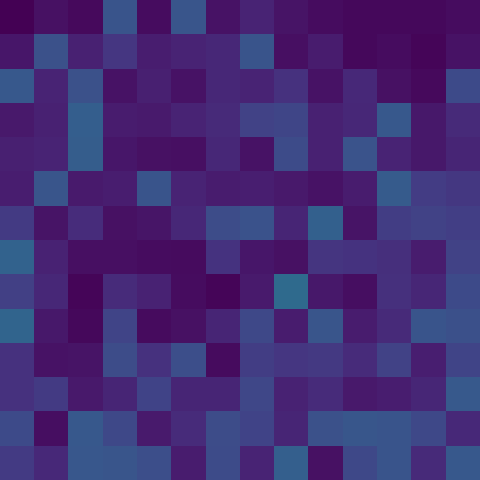}
\end{subfigure}
\begin{subfigure}{0.12\textwidth}
	\includegraphics[width=\textwidth]{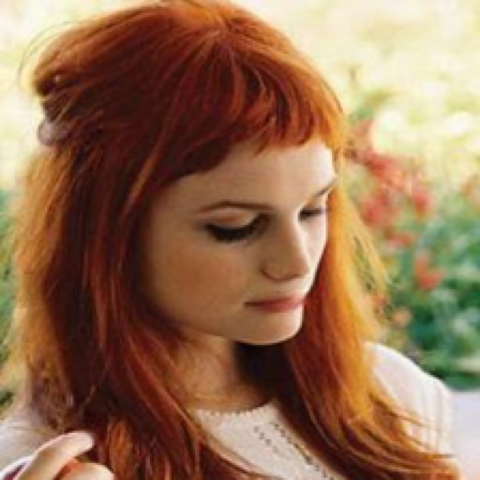}
\end{subfigure}
\VM

\begin{subfigure}{0.12\textwidth}
	\includegraphics[width=\textwidth]{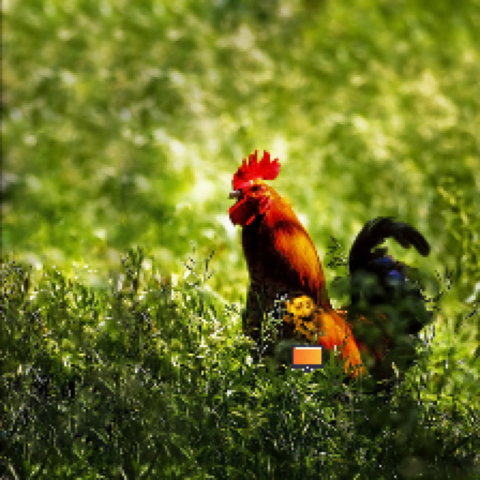}
\end{subfigure}
\begin{subfigure}{0.12\textwidth}
	\includegraphics[width=\textwidth]{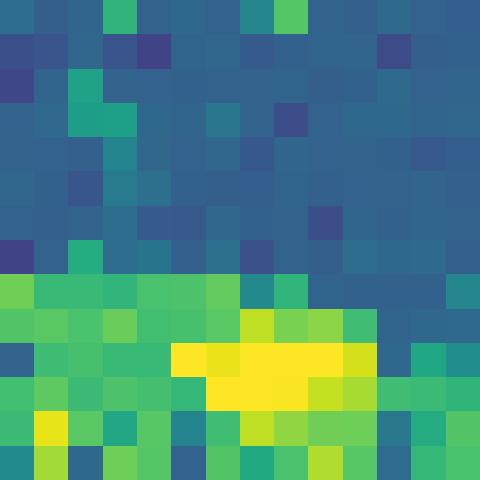}
\end{subfigure}
\begin{subfigure}{0.12\textwidth}
	\includegraphics[width=\textwidth]{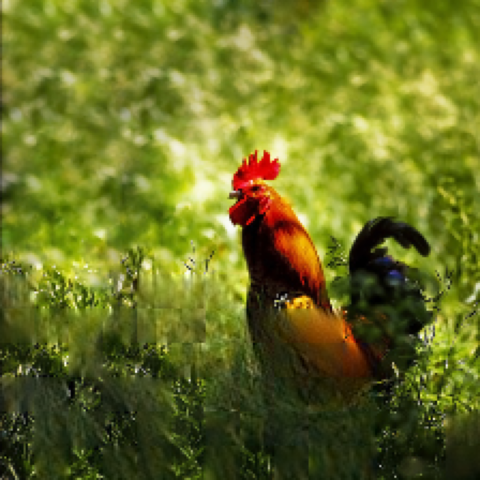}
\end{subfigure}
\begin{subfigure}{0.12\textwidth}
	\includegraphics[width=\textwidth]{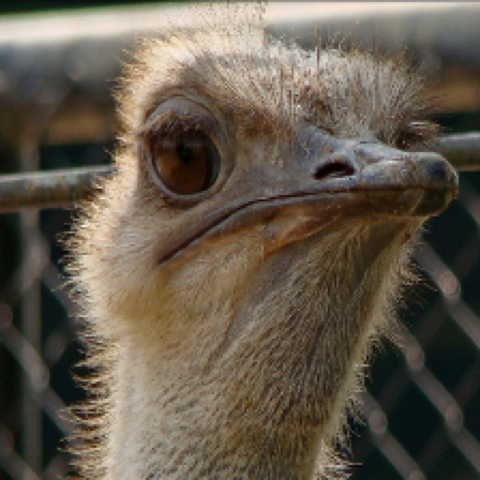}
\end{subfigure}
\begin{subfigure}{0.12\textwidth}
	\includegraphics[width=\textwidth]{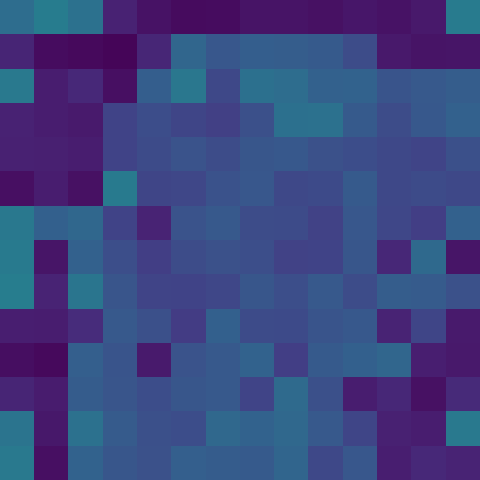}
\end{subfigure}
\begin{subfigure}{0.12\textwidth}
	\includegraphics[width=\textwidth]{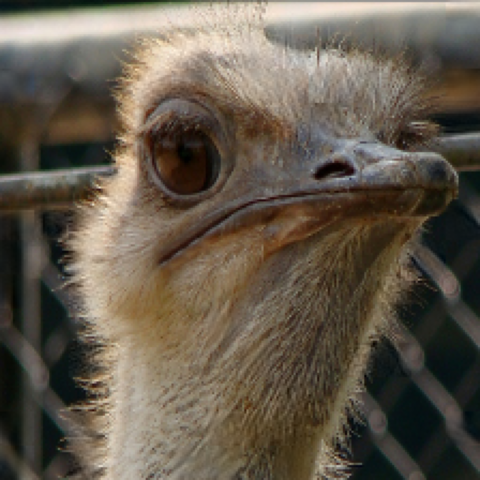}
\end{subfigure}
\VM

\begin{subfigure}{0.12\textwidth}
	\includegraphics[width=\textwidth]{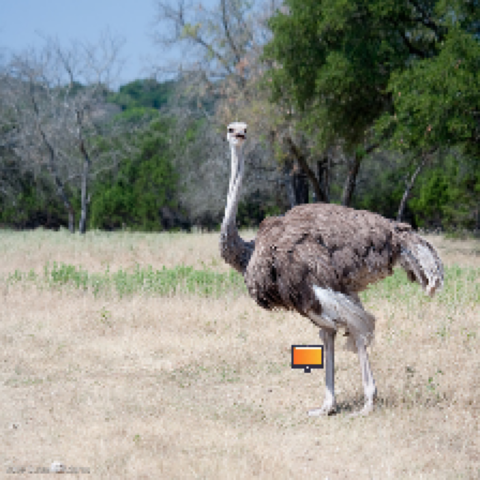}
\end{subfigure}
\begin{subfigure}{0.12\textwidth}
	\includegraphics[width=\textwidth]{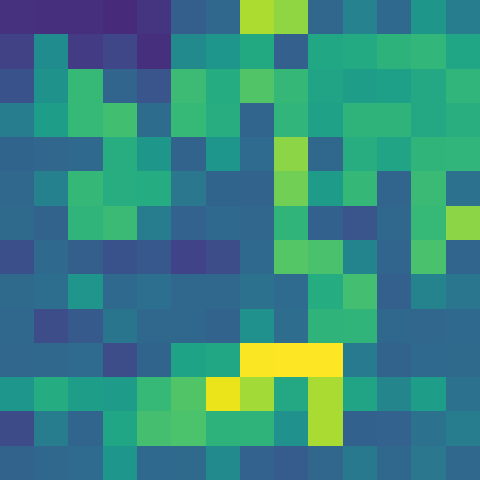}
\end{subfigure}
\begin{subfigure}{0.12\textwidth}
	\includegraphics[width=\textwidth]{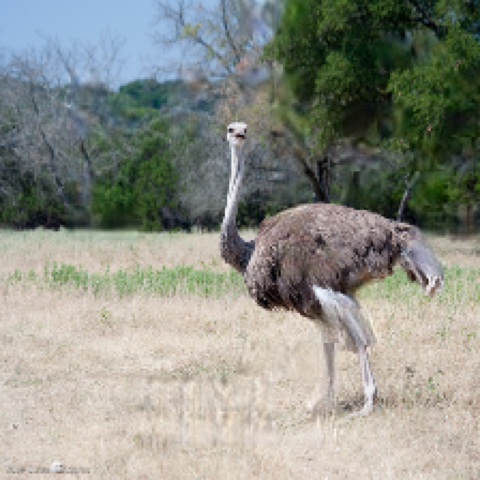}
\end{subfigure}
\begin{subfigure}{0.12\textwidth}
\includegraphics[width=\textwidth]{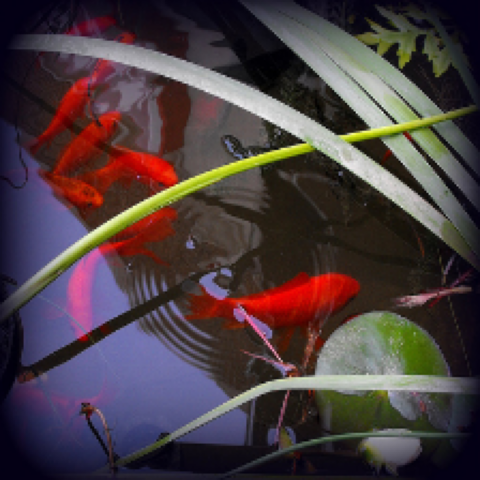}
\end{subfigure}
\begin{subfigure}{0.12\textwidth}
	\includegraphics[width=\textwidth]{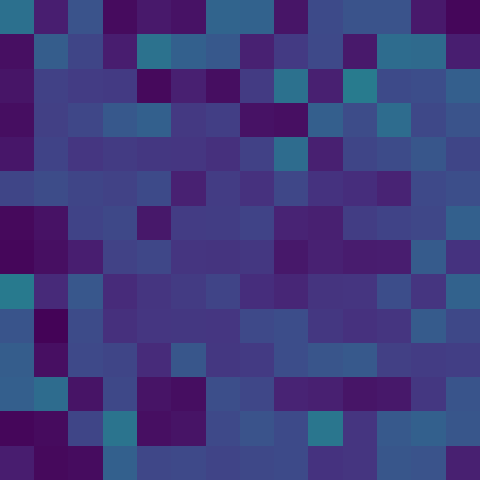}
\end{subfigure}
\begin{subfigure}{0.12\textwidth}
	\includegraphics[width=\textwidth]{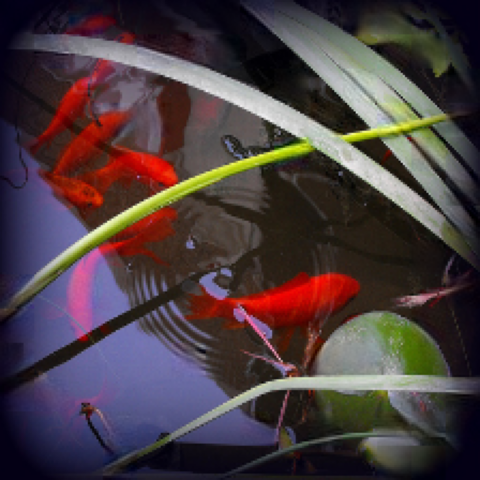}
\end{subfigure}

\begin{minipage}{0.12\textwidth}
	\centering
	\scriptsize Backdoored image 
\end{minipage} 
\begin{minipage}{0.12\textwidth}
	\centering
	\scriptsize $S$ 
\end{minipage} 
\begin{minipage}{0.12\textwidth}
	\centering
	\scriptsize Purified image 
\end{minipage} 
\begin{minipage}{0.12\textwidth}
	\centering
	\scriptsize Clean image 
\end{minipage} 
\begin{minipage}{0.12\textwidth}
	\centering
	\scriptsize $S$ 
\end{minipage} 
\begin{minipage}{0.12\textwidth}
	\centering
	\scriptsize Purified image 
\end{minipage}

\caption{Visualizations on backdoored / clean images.}
\label{fig:sup:clean_score}
\vspace{2mm}
\end{figure*}

\begin{table*}[!t]
\caption{Defense results on clean models.}
\renewcommand{\tabcolsep}{0.06cm}
\begin{center}
\scriptsize
 \scalebox{1.0}{
	\begin{tabular}{p{0.7cm}<{\centering}p{1.0cm}<{\centering}|p{0.76cm}<{\centering}p{0.76cm}<{\centering}|p{0.76cm}<{\centering}p{0.76cm}<{\centering}|p{0.76cm}<{\centering}p{0.76cm}<{\centering}|p{0.76cm}<{\centering}p{0.76cm}<{\centering}|p{0.76cm}<{\centering}p{0.76cm}<{\centering}|p{0.76cm}<{\centering}p{0.76cm}<{\centering}}
		\toprule
		\multirow{2}{*}{} & & \multicolumn{2}{c|}{\texttt{Cifar10}} & \multicolumn{2}{c|}{\texttt{GTSRB}} & \multicolumn{2}{c|}{\texttt{VGGFace2}} & \multicolumn{2}{c|}{\texttt{ImageNet10}} & \multicolumn{2}{c|}{\texttt{ImageNet50}} & \multicolumn{2}{c}{\texttt{ImageNet100}} \\
		\cmidrule{3-14}
		& & CA & BA & CA & BA & CA & BA & CA & BA & CA & BA & CA & BA \\
		\midrule
		\multicolumn{2}{c|}{Before Defense}  & 93.8 & 93.7 & 98.7 & 98.6 & 95.7 & 95.7 & 89.8 & 88.8 & 84.2 & 83.6 & 82.7 & 82.2\\
		\midrule
		\multirow{2}{*}{Ours} & Base   & 93.1 & 93.1 & 98.5 & 98.5 & 91.5 & 91.4 & 81.3 & 80.7 & 61.8 & 61.4 & 59.3 & 59.6\\
		& Large &  93.3 & 93.2 & 98.6 & 98.6 & 92.8 & 92.6 & 85.3 & 84.1 & 72.5 & 72.5 & 69.8 & 70.1\\
		\bottomrule
	\end{tabular} }
    \end{center}	
\vspace{0mm}
\label{tab:sup:clean-model}
\end{table*}

\begin{figure*}[!t]
 \centering	
 \includegraphics[width=\textwidth]{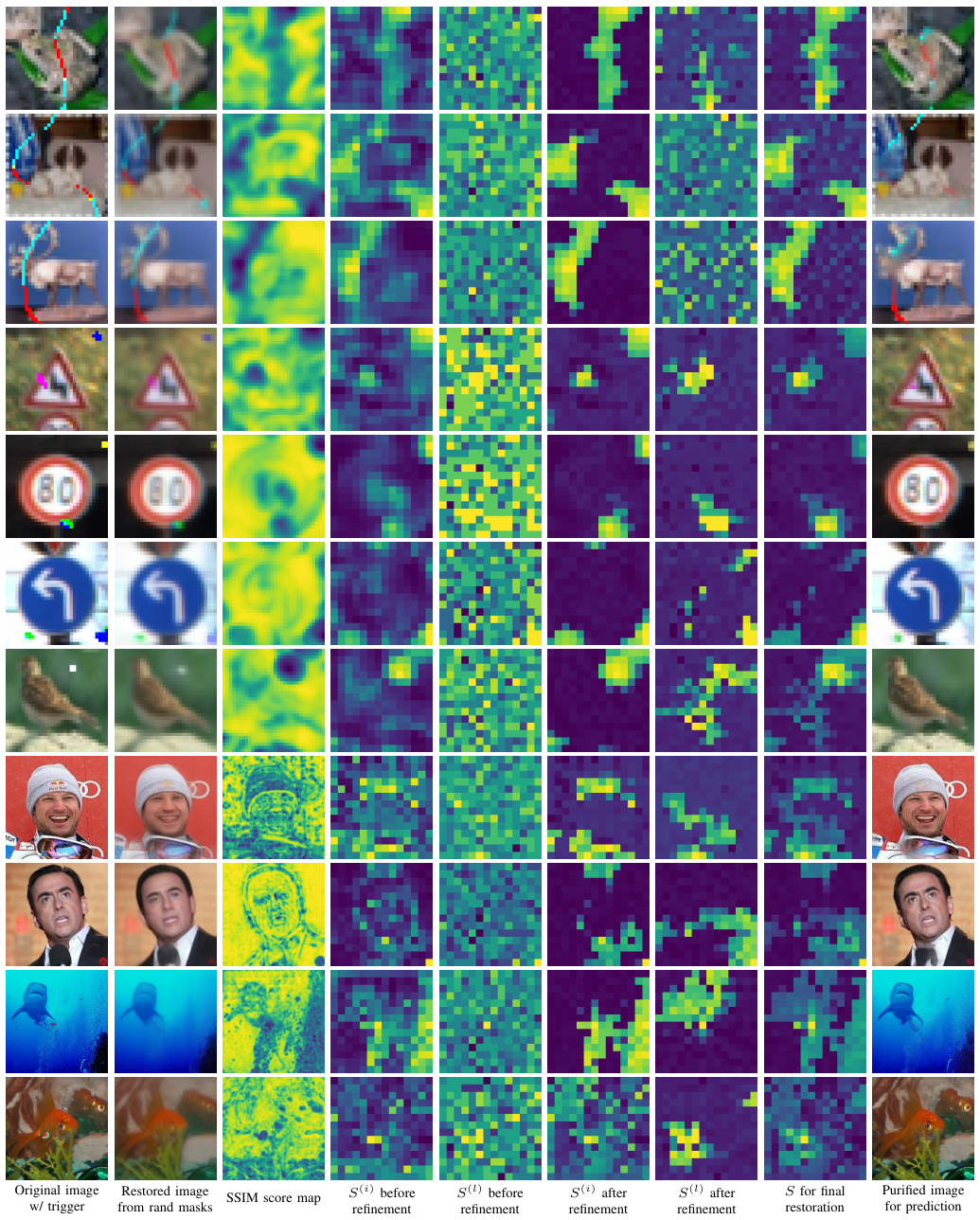} 	
    \caption{Additional sampled visualizations of the defense process. All the scores are clipped to a range of [0,1], with yellow for high values and blue for low values. The top six rows are from IAB attack, and the rest are from BadNet attack.}\label{fig:visualization_II}
\end{figure*}

\begin{figure*}[!t]
	\centering
	\includegraphics[width=\textwidth]{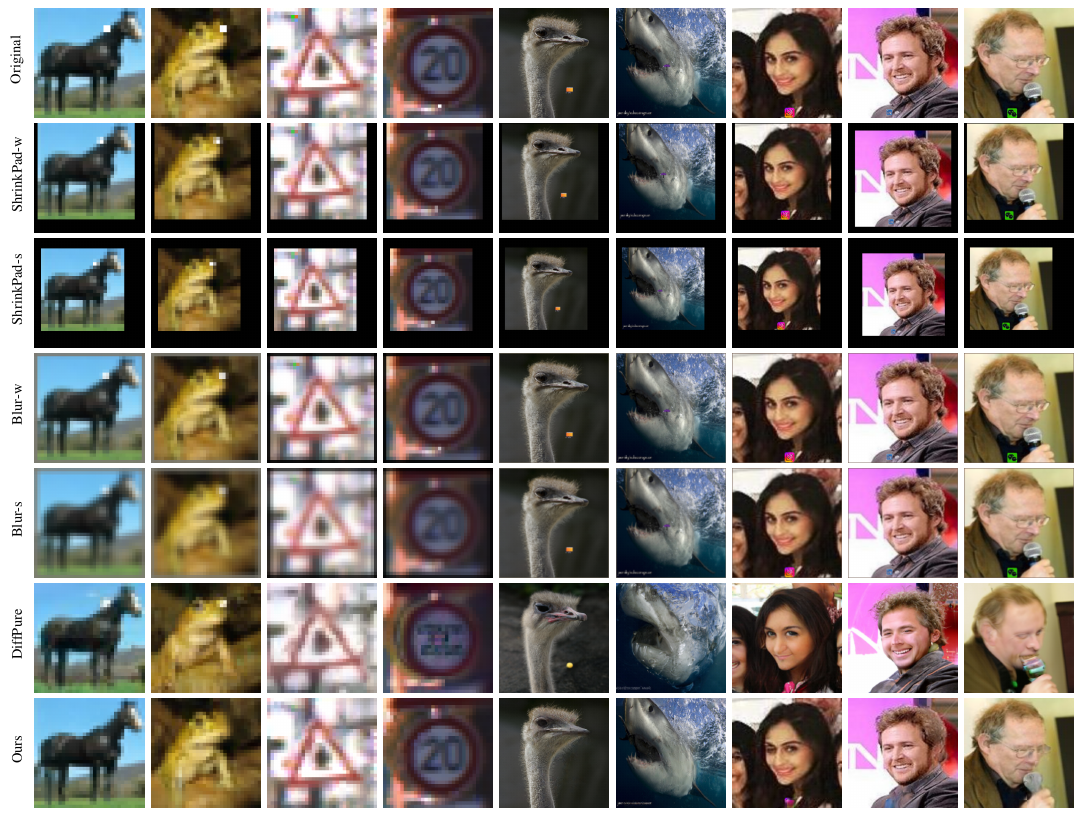}	
	\caption{Sampled visualizations of original images with triggers and images after defense (I).}\label{fig:vis_all_1}
 \vspace{2mm}
\end{figure*}

\begin{figure*}[!t]
	\centering	
      \includegraphics[width=\textwidth]{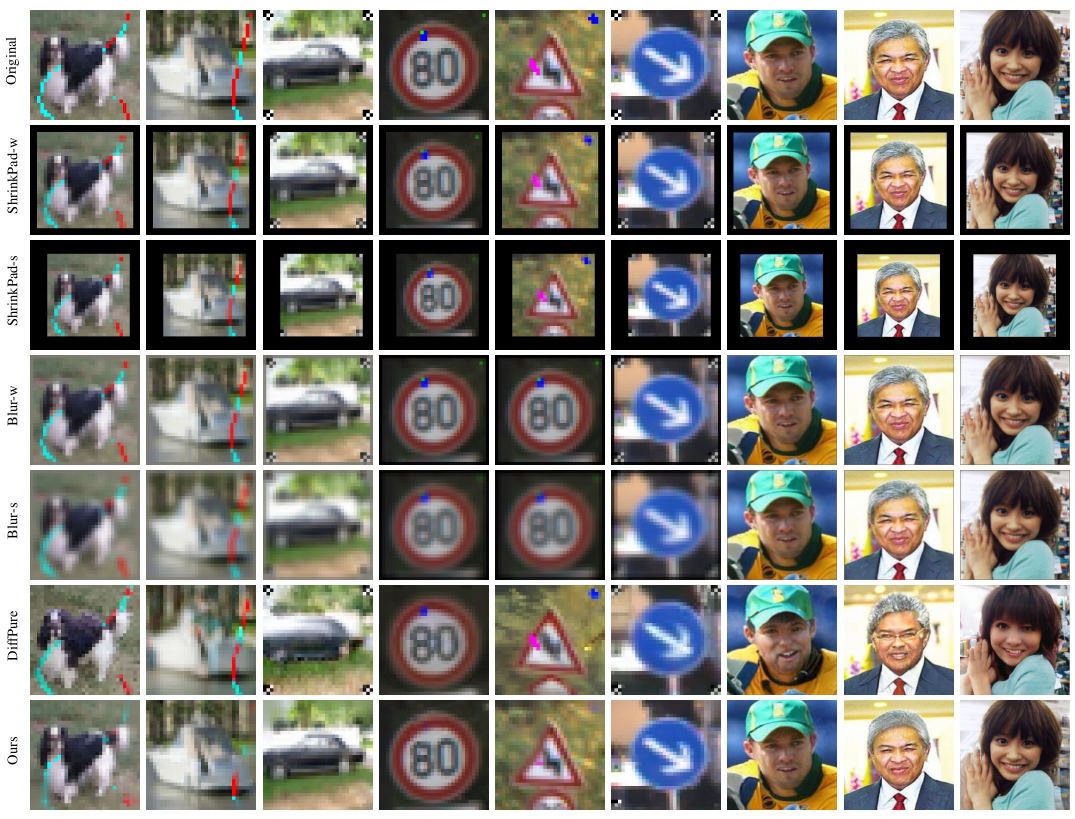}	
	\caption{Sampled visualizations of original images with triggers and images after defense (II).}\label{fig:vis_all_2}
 \vspace{2mm}
\end{figure*}

\subsection{Generalization on Network Architecture} \label{sec:arch}
Our method is for black-box defense, thus is generalizable on network architectures. In Tab.~\ref{tab:sup:arch}, we show results on \texttt{Cifar10} and \texttt{VGGFace2} with different backbone networks. Februus is a white-box method, thus it relies on the network architecture. On \texttt{Cifar10}, it performs well on the shallow Convolutional Neural Network originally used by the authors, but is less effective on ResNet18 and VGG16. On \texttt{VGGFace2}, its clean accuracies are relatively low. Compared with other black-box methods, our method achieves consistently better performances across different network architectures.

\subsection{Generalization to Clean Images and Clean Models}\label{sec:clean} 

We highlight that our method is blind to the benignity of images or models. This relies on dedicated designs in different stages. The key to guarantee correct label prediction in the situation of clean images or clean models is not destroying semantic contents in clean regions. Since the final prediction is based on MAE restoration $\rho(\x)$ from the final trigger score $S$, we should keep small score values of $S$ for those clean regions throughout the test-time defense process.

In the trigger score generation stage, if $\x$ is a clean image no matter $f$ is backdoored or not, its MAE restorations should be similar to the original image. This implies that values of $S^{(i)}$ will be small. The values of $S^{(l)}$ will also be small as the label prediction is unlikely to change. If $f$ is a clean model and $x$ is a backdoored image, $S^{(l)}$ will still be small. Although $S^{(i)}$ has high values for trigger region, its impact is reduced when we average $S^{(l)}$ and $S^{(i)}$. In the topology-aware score refinement stage, only the top $L$ tokens are affected. By construction $L=\sum_{r,c}\mathbb{I}[S^{(i)}_{r,c}\geq 0.2]$ or $L=\sum_{r,c} S^{(l)}_{r,c}$, $L$ is generally small in the situation of clean images or clean models. In the adaptive image restoration stage, image regions with trigger scores greater than $\tau_K$ are generated with MAE. These regions are either trigger regions or some content-irrelevant regions. The rest clean content regions are kept intact. Therefore, the model can still make correct label prediction on the purified image $\rho(\x)$.

For backdoored models on clean images, the CA in previous results has validated the effectiveness of our method. Figure~\ref{fig:sup:clean_score} shows different properties of trigger score $S$ between backdoored and clean images. $S$ of clean images has small values, thus the image restoration stage will not change the semantic content. For clean models, Table~\ref{tab:sup:clean-model} lists prediction accuracies for six different datasets. As can be seen, the accuracies on backdoored and clean images are minimally affected.

\subsection{Visualizations of Defense Process} We plot images and scores in Fig.~\ref{fig:visualization_I}. Restored images from random masks have the same content as the original images, but are different in the trigger regions and some details. This is reflected in the SSIM score map. The two trigger scores are slightly higher in the trigger region, but very noisy. After refinement, high scores concentrate on the triggers, and scores of content regions are suppressed. $S$ is then used to generate the purified images. Compared with the original backdoored images, triggers are removed while the image contents are preserved. The purified images lead to correct label predictions.

Additional visualization of the defense process are presented in Fig.~\ref{fig:visualization_II}. The top six rows come from IAB~\cite{nguyen2020IAB} attack. IAB uses sample-specific triggers, \ie, test images contain different triggers for one backdoored model. On \texttt{Cifar10}, the triggers are irregular curves. On \texttt{GTSRB}, the triggers are color patches. Due to the complexity of triggers, the heuristic search in image space using rectangle trigger blockers~\cite{udeshi2022model} may not work well. In our method, the refined trigger score $S$ successfully identifies the trigger in each test image. Triggers are removed in the purified images, leading to correct label predictions. On \texttt{VGGFace2} and \texttt{ImageNet10}, despite their larger image size, our method also manages to locate the tiny triggers and restore the clean images.

\subsection{Visualization of Defense Results}\label{sec:visualization_results}
Figures~\ref{fig:vis_all_1}-\ref{fig:vis_all_2} visualize the purified images of comparison defense methods. ShrinkPad and Blur apply global transformations on the images. They cannot remove the backdoor triggers, but sometimes can incapacitate the backdoor triggers through adding noises or distorting the trigger patterns. When the trigger patterns are large (\ie, IAB in Figure~\ref{fig:vis_all_2}), a strong transformation would be required to reduce ASR. But this will also sacrifice clean accuracies. 

DiffPure first adds a small amount noise to the backdoored images, and then uses a reserve generative process to recover the clean images. However, it frequently hallucinates image content. Looking at the last three columns of \texttt{VGGFace2}, DiffPure changes the facial expressions and facial features. These fine-grained attributes are critical to face recognition. For other datasets, DiffPure may not recover digits of \texttt{GTSRB} and the trigger patterns remain in the images. Although sometimes the trigger patterns are incapacitated. These visualization clearly shows the difference between our method and DiffPure, even though they both leverage large pretrained generative models. Ours only restores trigger-related regions, and keep other clean regions that contains important semantic details intact.

\subsection{Discussion and Limitation}
Test-time backdoor defense has drawn increasing attention. It is a practical yet challenging task. Only model predictions on the single test image can be used, while the backdoor attack methods can be quite diverse. By leveraging pretrained MAE models, our method locates possible triggers inside images and restores the missing contents simultaneously. We demonstrate its effectiveness on backdoor triggers of different patterns and shapes. One limitation of our method is that it focuses on the local attacks. Some particular attack methods use triggers that overlap the entire image. Since global triggers are less robust against image transformations, an additional transformation may be applied before our method~\cite{shi2023black}. 

\vspace{0mm}
\section{Conclusion}
\vspace{0mm}In this paper, we study the novel yet practical task of blind backdoor defense at test time, in particular for local attacks and black-box models. We devise a general framework of \textit{Blind Defense with Masked AutoEncoder} (BDMAE) to detect triggers and restore images. Extensive experiments on four benchmarks under various backdoor attack settings verify its effectiveness and generalizability.

\bibliographystyle{ieee}
\bibliography{reference}


 




\begin{IEEEbiography}[{\includegraphics[width=1in,height=1.25in,clip,keepaspectratio]{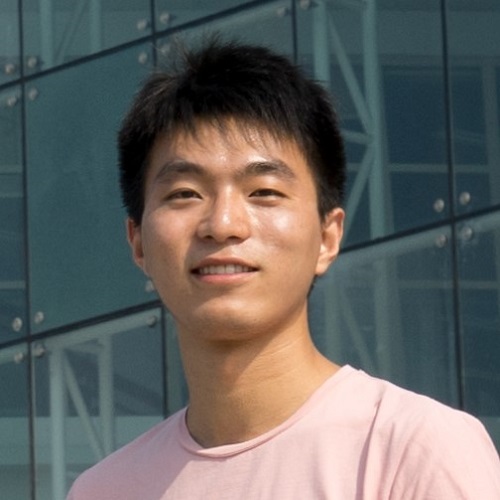}}]{Tao Sun}
  received the B.S. degree from Huazhong University of Science and Technology, Wuhan, China, in 2015, and M.S. degree from Nanjing University, Nanjing, China, in 2018. He is currently pursuing a Ph.D. degree in Computer Science at Stony Brook University. His current research interests include computer vision, and machine learning. 
\end{IEEEbiography}

\begin{IEEEbiography}[{\includegraphics[width=1in,height=1.25in,clip,keepaspectratio]{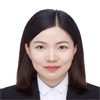}}]{Lu Pang}
  received the B.S and M.S from Peking University, Beijing, China. She is currently pursuing a Ph.D. degree in Computer Science at Stony Brook University, NY, USA. Her research interests include computer vision, machine learning security.
\end{IEEEbiography}

\begin{IEEEbiography}[{\includegraphics[width=1in,height=1.25in,clip,keepaspectratio]{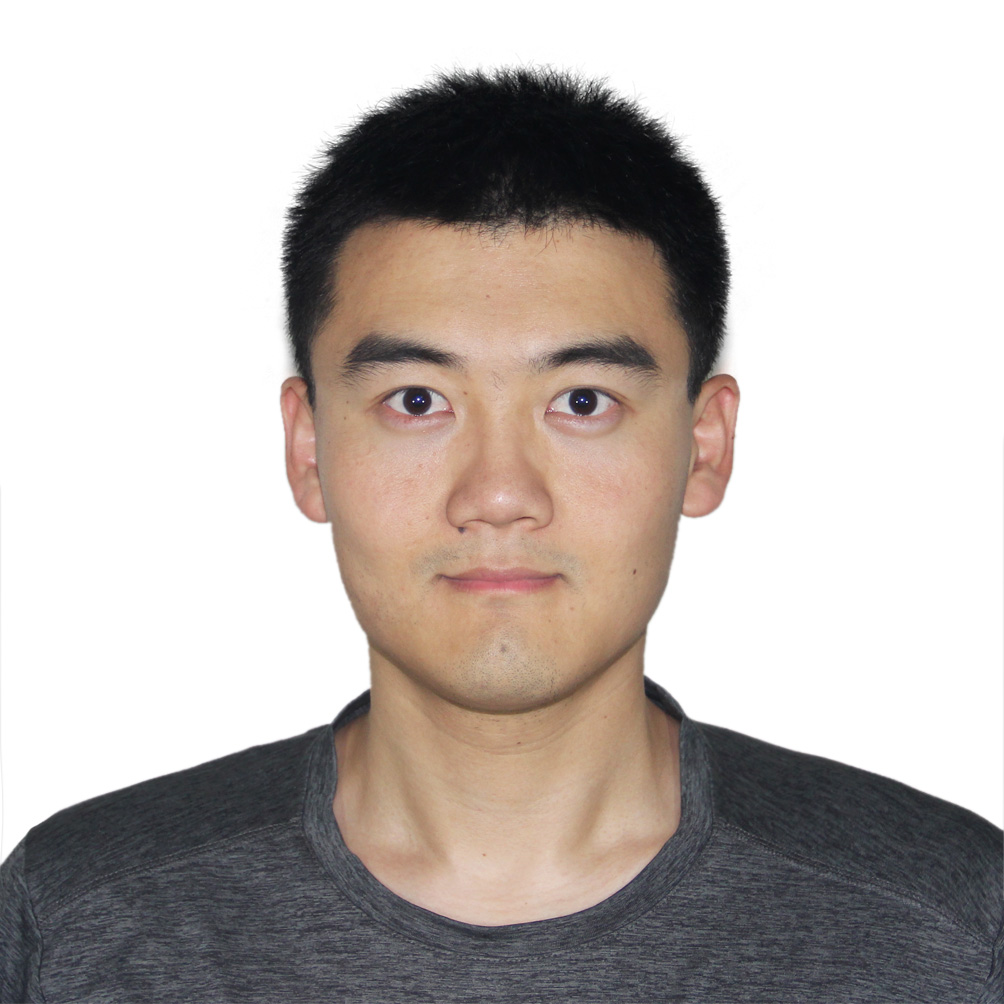}}]{Weimin Lyu}
  received the B.S. in Statistics from Shandong University, China. He is currently pursuing a Ph.D. degree in Computer Science at Stony Brook University, USA. His current research interests include ML security, Multimodal Large Language Model. 
\end{IEEEbiography}

\begin{IEEEbiography}[{\includegraphics[width=1in,height=1.25in,clip,keepaspectratio]{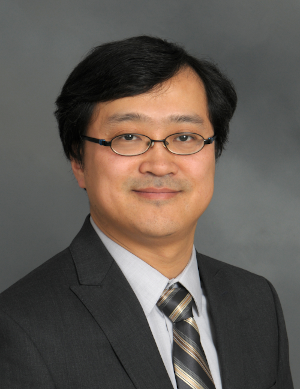}}]{Chao Chen} is an Associate Professor of Biomedical Informatics at the Stony Brook University, United States. He received his B.S from Peking University, and completed his Ph.D program at Rensselaer Polytechnic Institute. His research interests include machine learning, biomedical image analysis, and topological data analysis.
\end{IEEEbiography}

\begin{IEEEbiography}[{\includegraphics[width=1in,height=1.25in,clip,keepaspectratio]{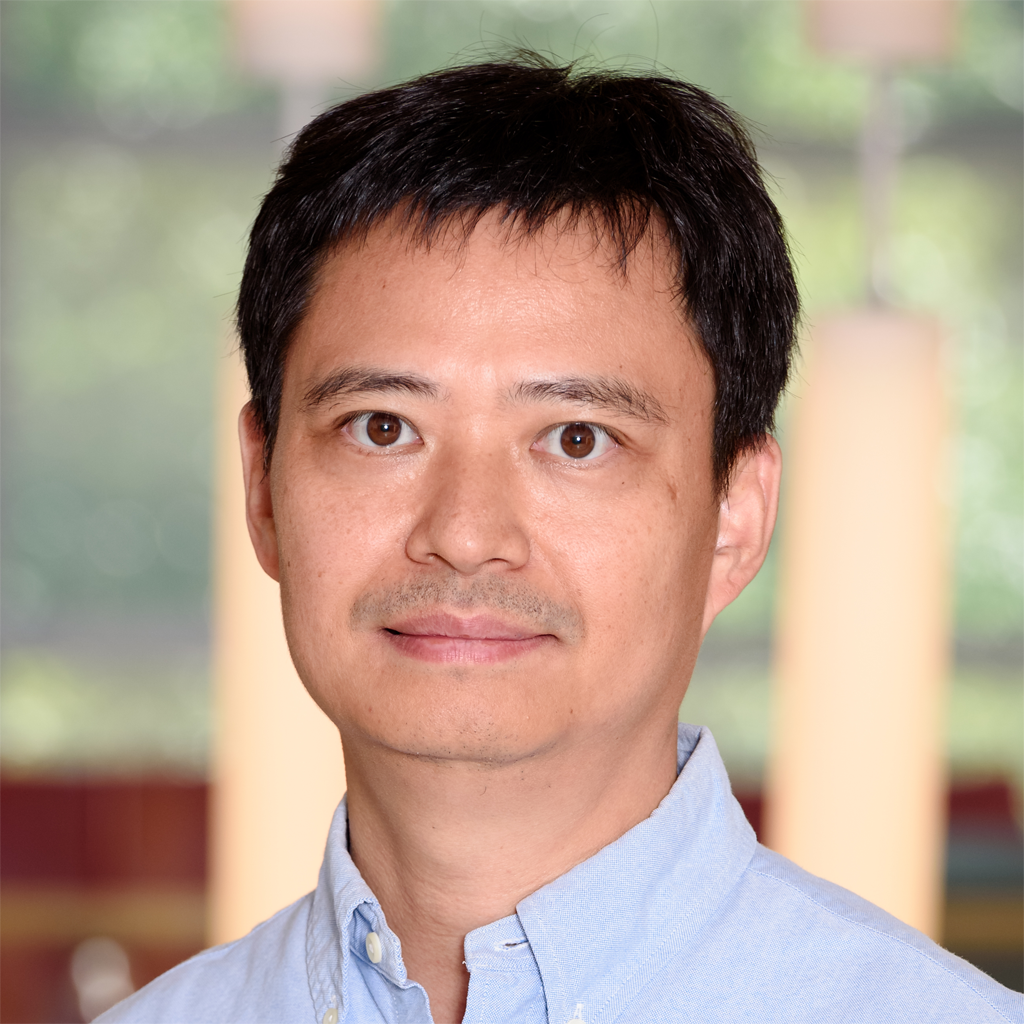}}]{Haibin Ling}
 Haibin Ling received B.S. and M.S. from Peking University in 1997 and 2000, respectively, and Ph.D. from University of Maryland in 2006. From 2000 to 2001, he was an assistant researcher at Microsoft Research Asia; from 2006 to 2007, he worked as a postdoctoral scientist at UCLA; from 2007 to 2008, he worked for Siemens Corporate Research as a research scientist; and from 2008 to 2019, he was first an assistant and then an associate professor of Temple University. In fall 2019, he joined the Department of Computer Science of Stony Brook University, where he is now a SUNY Empire Innovation Professor. His research interests include computer vision, augmented reality, medical image analysis, visual privacy protection, and human computer interaction. He received Best Student Paper Award of ACM UIST (2003), Best Journal Paper Award at IEEE VR (2021), NSF CAREER Award (2014), Yahoo Faculty Research and Engagement Award (2019), and Amazon Machine Learning Research Award (2019). He serves or served as associate editors for IEEE Trans. on Pattern Analysis and Machine Intelligence (PAMI), IEEE Trans. on Visualization and Computer Graphics (TVCG), Pattern Recognition (PR), and Computer Vision and Image Understanding (CVIU), and as Area Chairs for major computer vision relevant conferences such as CVPR, ICCV, ECCV, ACM MM, and WACV. He is a Fellow of IEEE.
\end{IEEEbiography}

\vfill

\end{document}